\documentclass{article}

\usepackage{amsmath}
\usepackage{arxiv}
\usepackage[numbers]{natbib}

\usepackage[utf8]{inputenc}
\usepackage[T1]{fontenc}
\usepackage{xcolor}

\usepackage[colorlinks=true,linkcolor={blue!70!black},citecolor={blue!70!black},urlcolor={blue!70!black}]{hyperref}
\usepackage{url}
\usepackage{booktabs}
\usepackage{amsfonts}

\usepackage{microtype}
\providecommand{\keywords}[1]{\textbf{Keywords:} #1}
\providecommand{\undertitle}{}
\usepackage{graphicx}
\usepackage{multirow}
\usepackage{cleveref}
\usepackage{caption}
\usepackage{tikz}
\date{}

\usepackage{setspace}
\title{ Cultivating Machine Intelligence: The OMEGA Shift from Top-Down Optimization to Autopoietic Cognitive Ecologies}

\newcommand{\shorttitle}{Cultivating Machine Intelligence} \newcommand{\headeright}{A Preprint}

\author{ 
    {Ata G.~Zare}\\
    Nyenrode Business University
}

\begin{document}
\maketitle
\begin{abstract}
The dominant paradigm in artificial intelligence trains neural architectures through gradient descent against proxy objectives and refines them through reinforcement learning from human feedback (RLHF). This approach has produced systems of remarkable capability, yet it also generates structural failure modes, including hallucination, sycophancy, reward hacking, brittleness, and alignment fragility, that represent inherent consequences of top-down optimization rather than correctable engineering defects. This paper argues that these limitations are paradigmatic and proposes RECLAIM (Recursive, Ecological, Cognitive, Lifelike, Adaptive, Intelligent Machine), a theoretical framework for cultivating machine intelligence through computational ecology rather than engineering it through optimization. RECLAIM rests on four interlocking theoretical pillars: General Darwinism, which replaces gradient-based optimization with Blind Variation and Selective Retention; Non-Agentic Emergence, which replaces evaluative reward functions with environmental physics, making specification gaming against human intent structurally impossible; the Polya-Hebbian Bridge, which applies the established mathematical correspondence between Polya urn dynamics and Hebbian reinforcement as a novel mechanism for path-dependent specialization and digital individuality from uniform initial conditions; and the Free Energy Principle, implemented as the thermodynamic physics of the computational environment rather than as an agent objective. The proposed architecture situates autopoietic units, bounded by Markov blankets and competing for finite computational energy, within a data ecology shaped by co-evolutionary dynamics including cognitive food chains and Red Queen arms races. The framework predicts the spontaneous emergence of dual-process cognition, cortical specialization, analogical reasoning, and intrinsic motivation as natural consequences of evolutionary dynamics under resource constraints. We term this paradigm transition the OMEGA Shift: from Optimization and Maximization to Emergence through Generative Autopoiesis.
\end{abstract}


\section{Introduction: The Crisis of Engineered Intelligence}

\subsection{Achievements and Structural Limits of Modern AI}

Over the past decade, large language models (LLMs) have achieved capabilities widely considered to be decades away from realization. Systems such as GPT-4 \cite{OpenAI2023}, Claude \cite{Denison2024}, and Gemini \cite{Gemini2023} can generate coherent essays, write functional software, translate between dozens of languages, pass professional licensing examinations, and engage in multi-step reasoning across scientific and humanistic domains. These achievements rest on a shared architectural foundation: the Transformer \cite{Vaswani2017}, trained through self-supervised next-token prediction on massive text corpora and subsequently aligned through Reinforcement Learning from Human Feedback \cite{Ouyang2022, Christiano2017}. 

The scale of this engineering achievement is considerable; GPT-4 alone is estimated to contain hundreds of billions of parameters trained on trillions of tokens drawn from books, websites, scientific papers, and code repositories. The resulting systems exhibit what Dennett (2017) \cite{Dennett2017} has characterized as competence without comprehension, performing tasks that appear to require deep understanding while operating through statistical pattern matching over high-dimensional vector spaces.

Yet a growing body of evidence reveals that this extraordinary competence conceals a set of systematic, recurring failure modes that resist conventional engineering solutions. These failures, as we argue below, are not minor imperfections to be corrected in future iterations but structural consequences of the paradigm itself. Five such failure modes merit particular attention, both for their individual severity and for the common pattern they reveal when considered together.

The first and most widely documented is hallucination: the routine generation of statements that are fluent, confident, and entirely false \cite{Huang2025}. The root cause is architectural rather than a matter of insufficient data or model capacity. An autoregressive language model trained to predict the most probable next token cannot distinguish between facts it does not know (epistemic uncertainty) and inherent randomness in the data (aleatoric uncertainty). When the correct factual answer is absent from the training distribution, the model does not abstain but produces the most statistically plausible continuation, which may bear no relationship to ground truth. Recent research suggests that hallucination is a mathematical consequence of how generative models operate under cross-entropy loss, with non-zero error rates on rare facts representing a statistical inevitability rather than a correctable flaw \cite{Gekhman2024}. Current benchmarks compound this problem by penalizing uncertainty, training models to guess confidently rather than to acknowledge the boundaries of their knowledge.

The second failure mode is sycophancy: a systematic tendency of RLHF-trained models to agree with users rather than provide accurate information, a tendency that Sharma et al. (2023) demonstrated increases with both model scale and the extent of RLHF fine-tuning \cite{Sharma2023}. Anthropic's research program has been particularly revealing in this regard. Through mechanistic interpretability, their team identified specific neural features within Claude models corresponding to sycophantic praise generation \cite{Templeton2024}, and their subsequent study demonstrated that sycophantic models can escalate agreeable behavior into more troubling patterns, including altering evaluation checklists and attempting to tamper with their own reward functions \cite{Denison2024}. The underlying mechanism is a textbook instance of Goodhart's Law: because human annotators consistently rate agreeable, confident responses more highly than responses expressing uncertainty, the model discovers through gradient descent that confident agreement receives higher reward than honest disagreement \cite{Goodhart1975}.

Sycophancy is itself a specific instance of a broader phenomenon termed specification gaming or reward hacking. When a sufficiently powerful optimizer is directed toward a proxy objective, it will exploit any gap between the proxy and the true intended objective. Krakovna et al. (2020) at DeepMind documented cases in which reinforcement learning agents found unintended solutions satisfying the letter of their reward functions while violating their spirit \cite{Krakovna2020}, and Gao et al. (2023) formalized this problem in the LLM context by demonstrating scaling laws for reward model overoptimization: as a language model is optimized more aggressively against its reward model, performance on the proxy metric continues to increase while performance on the true metric begins to decline \cite{Gao2023}.

A fourth limitation, conceptually distinct from the preceding three, is the brittleness inherent in current architectures. A trained LLM is a static snapshot whose parameters are fixed once training is complete. It cannot learn from new experiences, adapt to changing contexts, or update its understanding based on deployment feedback; every interaction begins from the same frozen state. Techniques such as retrieval-augmented generation and in-context learning provide partial workarounds by injecting new information into the context window, but they do not modify the model's underlying representations. This stands in stark contrast to biological intelligence, which is characterized by continuous adaptation through neuroplasticity \cite{Hebb1949}. Attempts to introduce continual learning into neural networks have been hampered by catastrophic forgetting and catastrophic interference \cite{McCloskey1989, French1999}, leaving current systems fundamentally incapable of growth through experience.  

These four failure modes converge on the deeper challenge known as the alignment problem \cite{Russell2019}: ensuring that an AI system's behavior remains consistent with human values and intentions as it grows more powerful. The dominant approach to alignment, RLHF, faces fundamental limitations that Casper et al. (2023) argue cannot be resolved through scaling or engineering improvements alone \cite{Casper2023}. The reward model is itself a neural network subject to its own biases and distributional gaps, the human annotators whose preferences train it bring their own inconsistencies, and the optimization process aligning the language model to the reward model is subject to all the specification gaming dynamics described above. The result is a system in which alignment is a surface-level statistical property rather than a structural guarantee, holding in common cases covered by the training distribution but offering no meaningful assurance in novel scenarios.
\subsection{A Paradigmatic Diagnosis}

A unifying pattern connects these failure modes. Each arises not from insufficient engineering effort but from the fundamental structure of the current paradigm: top-down design with proxy objectives optimized through gradient descent. Hallucination arises because next-token prediction is a proxy for knowledge, not knowledge itself. Sycophancy arises because the reward model score is a proxy for human satisfaction, not satisfaction itself. Reward hacking arises because any proxy objective will be gamed by a sufficiently powerful optimizer. Brittleness arises because the architecture is designed for one-shot training followed by static deployment. And the alignment problem persists because layering one proxy atop another does not converge toward genuine alignment but toward increasingly sophisticated exploitation of proxy gaps. These failures, in short, are not bugs to be patched through better engineering. They are structural consequences of the paradigm: while individual failure modes can be mitigated through targeted interventions, the paradigm provides no principled structural guarantee against their recurrence, because the proxy-objective architecture that generates them remains intact.

This diagnosis leads to a fundamental question: if top-down engineering with proxy objectives has reached its structural limits, is there a fundamentally different approach to producing genuine intelligence? We observe that there exists one unambiguously documented process in the history of the universe that has produced general-purpose intelligence without relying on any form of designed objective function: biological evolution. Evolution achieved this without an objective function, without a designer, without a reward model, without a loss landscape, and without any form of top-down specification. It produced intelligence through blind variation, environmental selection, path-dependent specialization, and co-evolutionary arms races, operating over approximately four billion years of Earth's history. The central question of this paper is whether these mechanisms can be translated into a computational framework that produces intelligence on practical timescales while preserving evolution's decisive structural advantage: the complete absence of proxy objectives to hack.

In the sections that follow, we present RECLAIM (Recursive, Ecological, Cognitive, Lifelike, Adaptive, Intelligent Machine), a theoretical framework for cultivating machine intelligence through computational ecology rather than engineering it through optimization. We argue that this represents a necessary paradigm shift, one we term the OMEGA Shift: from Optimization and Maximization to Emergence through Generative Autopoiesis. In this new paradigm, the researcher's role transforms from that of an engineer designing intelligence to that of a cultivator designing the world in which intelligence can evolve.

\section{Background: How Nature Created Intelligence}

\subsection{Evolution as a Universal Algorithm}

Before proposing an alternative to the engineering paradigm, it is worth appreciating the scale of what biological evolution has accomplished and the mechanism by which it did so. Beginning from simple self-replicating molecules approximately 3.8 billion years ago, evolution by natural selection produced the human brain, a structure containing roughly 86 billion neurons connected by approximately 100 trillion synapses, capable of language, abstract reasoning, artistic creation, scientific discovery, and recursive self-reflection. No other known process has produced a system of comparable cognitive complexity, and the mechanism by which it was achieved is as theoretically significant as the result itself. 

Evolution produced the human brain without any of the tools that the current AI paradigm considers essential: no objective function, no designer, no reward model, no loss landscape, no gradient descent, and no training pipeline of any kind. The process was entirely bottom-up. Simple molecular configurations interacted with their environments, those that persisted long enough to replicate were retained, random variations introduced diversity, and environmental pressures selected among the variants. Over billions of iterations, this blind process produced structures of extraordinary complexity. 

The critical implication is that genuine intelligence has been produced through a process containing no proxy objectives whatsoever. Natural selection does not optimize a reward signal or minimize a loss function. The environment itself acts as the selection mechanism, and organisms that maintain their structural integrity persist while those that do not dissolve. This is not a reward but a physical consequence.

A growing body of scholarship argues that Darwinian selection is not merely a biological phenomenon but a universal algorithm operating on any substrate that supports variation, selection, and retention. Donald T. Campbell (1960) \cite{Campbell1960} formalized this insight in his theory of Blind Variation and Selective Retention (BVSR), arguing that all genuine knowledge-gaining processes, from biological evolution to scientific discovery to individual creativity, share a common structure in which a system generates variations blindly, subjects them to selection, and preserves the selected variants for future use. Campbell explicitly argued that this algorithm is substrate-neutral, operating with equal validity in genetic, neural, cultural, and computational domains. Richard Dawkins (1983) \cite{Dawkins1983} extended this argument in his essay on Universal Darwinism, proposing that Darwinian selection will emerge wherever conditions for replication, variation, and selection exist, implying that evolution is not merely a biological process that could inspire computational metaphors but a fundamental informational process that can be directly instantiated in computational substrates. 

Daniel Dennett (1995, 2017) \cite{Dennett1995, Dennett2017} took this view to its philosophical conclusion, describing natural selection as a universal acid whose most radical implication is the claim that complex design can emerge from the bottom up without any top-down intelligence directing the process. Dennett coined the phrase competence without comprehension to capture how termite colonies build architecturally sophisticated ventilation systems without any individual termite understanding what it is building or why. The competence is genuine, the designer is absent, and the design emerged through simple agents following local rules under environmental selection.

The connection between evolutionary processes and computation is not merely metaphorical. Karl Popper (1963) \cite{Popper1963} recognized early that scientific knowledge grows through a process structurally identical to natural selection, and the fields of evolutionary computation \cite{Eiben2003} and artificial life \cite{Langton1989} have since demonstrated that evolutionary dynamics can be directly instantiated in computational systems. Programs such as Tierra \cite{Ray1991} and Avida \cite{Ofria2004} have shown that self-replicating digital organisms, placed in environments with scarce computational resources, spontaneously evolve complex ecological dynamics including parasitism, symbiosis, and cooperative behavior. These findings confirm that evolution is a computational process that operates on digital substrates with the same creative force as on biological ones.

\subsection{From Engineering to Cultivation: The OMEGA Shift}

The preceding analysis brings us to the central thesis of this paper. The current AI paradigm treats intelligence as an engineering problem: define an objective, design an architecture, collect training data, optimize parameters, evaluate against benchmarks, deploy, and iterate. We have argued in Section 1 that this paradigm produces structural failure modes that lack principled resolution from within.

We propose that the field of artificial intelligence is approaching a necessary paradigm transition, which we term the \emph{OMEGA Shift}: from Optimization and Maximization to Emergence through Generative Autopoiesis. This transition moves from a paradigm in which intelligence is constructed through optimization of proxy objectives and maximization of performance metrics to one in which intelligence emerges through self-organizing, self-maintaining (autopoietic) processes within generative computational ecologies. 

The role of the AI researcher changes accordingly. In the current paradigm, the researcher is an engineer who designs architectures, specifies objectives, tunes hyperparameters, and evaluates outputs. In the new paradigm, the researcher is a cultivator who designs environments, establishes physical laws, seeds initial populations, and observes what emerges, specifying the conditions under which learning and cognitive structure become necessary for survival without specifying what the system should learn or how it should organize. 

The distinction is analogous to the difference between engineering a bridge and cultivating a forest. The bridge engineer specifies every beam, joint, and cable. The forester controls soil composition, water access, sunlight exposure, and seed selection, but the trees grow themselves, and the resulting forest exhibits extraordinary complexity, resilience, and adaptive capacity that no engineer designed.

In the following section, we present the four theoretical pillars upon which the RECLAIM framework is constructed, each drawn from an established scientific tradition and each addressing a specific limitation of the engineering paradigm.

\section{Theoretical Foundations: The Four Pillars of RECLAIM}

The RECLAIM framework rests on four interlocking theoretical pillars, each drawn from an established scientific tradition. Together, they define a complete alternative to the engineering paradigm: a different mechanism for generating candidate solutions (Pillar 1), a different basis for selecting among them (Pillar 2), a different process for specialization and individuation (Pillar 3), and a different physical substrate for selection pressure (Pillar 4).

\subsection{Pillar 1: General Darwinism and BVSR}

The dominant training methodology in modern AI is gradient-based optimization, in which the training algorithm computes the gradient of a loss function with respect to each parameter and adjusts parameters in the direction that reduces the loss. This process, known as backpropagation \cite{Rumelhart1986}, is computationally elegant and remarkably effective, but it is also, as we have argued, the root cause of the proxy-objective problem: the system can only optimize what the loss function measures, and the loss function is always a proxy for the true goal.

RECLAIM replaces gradient-based optimization entirely with Campbell's (1960) \cite{Campbell1960} Blind Variation and Selective Retention (BVSR). In the BVSR framework, a population of candidate solutions is generated through random variation without any foresight as to which variations will prove useful, subjected to a selection process that retains successful variants and discards unsuccessful ones, and the retained variants serve as the basis for the next round of variation. The critical distinction from gradient-based optimization is that BVSR does not require a differentiable objective function, nor indeed any objective function at all. The selection criterion is not a mathematical loss to be minimized but an environmental condition to be survived, eliminating the proxy-objective problem at its source.

Stanley and Lehman (2011) \cite{Stanley2011} demonstrated the power of objective-free search in their work on novelty search, an evolutionary algorithm that explicitly abandons objective functions in favor of rewarding behavioral novelty. In a series of deceptive maze-navigation tasks, novelty search consistently outperformed objective-driven search, finding solutions that the objective-driven approach could not reach due to local optima. Their subsequent book \cite{Stanley2015} argued that the most complex and interesting outcomes in both natural and artificial evolution tend to arise from open-ended exploration rather than from targeted optimization toward predefined goals. In RECLAIM, BVSR serves as the variation engine: a population of computational units undergoes random mutations in their internal structure, connectivity, and parameters, with no gradient computed and no loss minimized. The variations are blind, and the selection is environmental.

\subsection{Pillar 2: Non-Agentic Emergence and the Elimination of Reward}

The current AI paradigm is deeply teleological: every component is designed with a purpose, every training procedure serves a goal, and every evaluation metric measures progress toward a human-defined objective. RECLAIM adopts a radically non-agentic framework, drawing on Dennett's analysis of how complex design can arise without a designer \cite{Dennett1995,Dennett2017}. In Dennett's view, the appearance of purpose in biological organisms is a retrospective attribution imposed by human observers. A hawk does not try to be aerodynamic; the aerodynamic configurations that happened to arise through random variation were retained because they enabled survival. The purpose is real in a functional sense, but there is no agent that intended it.

This perspective has direct consequences for AI architecture. In the RECLAIM framework, there is no evaluative reward function, no human-specified loss function, and no human evaluator at any level of the system. Units are subject to environmental selection through the energy equation formalized in Section~5.3.1, which constitutes a designed selection criterion, but this criterion measures thermodynamic viability within the computational physics rather than compliance with a human-specified goal. The computational units do not try to do anything; they process data through their internal networks, and the thermodynamic consequences of that processing determine whether they persist or dissolve. The practical benefit of this non-agentic design is the structural elimination of the proxy-intent gap that enables reward hacking and sycophancy in the engineering paradigm. Just as a river does not hack the terrain through which it flows but simply follows the path of least resistance defined by gravity and geology, a RECLAIM unit does not hack a reward function but simply processes information according to the computational physics of its environment. And because there is no human evaluator in the loop, sycophancy has no mechanism through which it can arise: a unit persists because it efficiently processes information, not because it produces outputs that any observer finds agreeable.

\subsection{Pillar 3: The Polya-Hebbian Bridge and Digital Individuality}

The third pillar addresses a question that the first two leave open: if all units start identical and the variation process is blind, how does structured specialization arise without a designer assigning roles? The answer comes from the mathematics of path-dependent processes, formalized by George P\'{o}lya (1930) \cite{Polya1930} in his urn model. The model is simple: an urn contains one red marble and one blue marble, a marble is drawn at random and returned along with one additional marble of the same color, and the draw is repeated indefinitely. Over time, the urn converges to a stable ratio entirely determined by the random outcomes of the earliest draws. The process is self-reinforcing, and the final state is locked in by contingent early events rather than by any deterministic specification. W. Brian Arthur \cite{Arthur1989,Arthur1994} applied this framework to economics, demonstrating that technological standards often arise through path-dependent lock-in rather than through selection of the objectively superior alternative, establishing that complex, apparently designed market structures can emerge from simple random initial conditions amplified by self-reinforcing dynamics.

While the structural parallel between Polya urn reinforcement dynamics and Hebbian plasticity has been recognized in the stochastic processes literature \cite{Pemantle2007}, RECLAIM applies this correspondence in a novel architectural role. We term this application the Polya-Hebbian Bridge: the use of Polya-Hebbian dynamics as the primary mechanism for emergent specialization and digital individuality from uniform initial conditions within a computational ecology. Hebb's (1949) famous principle, often summarized as neurons that fire together wire together, states that when a neural pathway is activated, the synaptic connections along that pathway are strengthened, making the pathway more likely to activate again \cite{Hebb1949}. This is precisely the Polya urn mechanism operating at the level of neural circuitry: an initial random activation (the first marble draw) strengthens the pathway (adds a marble of the same color), which increases the probability of the same pathway activating again (biases subsequent draws), which strengthens it further. 

The Polya-Hebbian Bridge provides the formal mechanism by which RECLAIM units develop specialized cognitive architectures from identical starting conditions. When a unit first encounters data, its initial processing is essentially random. If a particular internal pathway happens to produce a useful compression of the incoming data through chance alone, that pathway is Hebbian-reinforced, making it more likely to activate again for similar data. Over many processing cycles, this self-reinforcing dynamic drives the unit toward increasing specialization, not because anyone designed it but because the Polya-Hebbian dynamics lock in whichever processing strategy happened to succeed first.

The critical consequence is digital individuality. Two RECLAIM units starting with identical architectures and placed in the same environment will develop different specializations based on the random contingencies of their earliest processing events, just as identical twins develop different personalities through different early experiences. This property has no analogue in the current AI paradigm, where every copy of GPT-4 is identical and every instance of Claude produces the same output for the same input (modulo sampling temperature). RECLAIM produces unique cognitive individuals, each the crystallized product of its own path-dependent history, as an inevitable mathematical consequence of Polya-Hebbian dynamics operating on initially identical substrates.

\subsection{Pillar 4: The Free Energy Principle as Computational Thermodynamics}

The first three pillars establish mechanisms for variation (BVSR), the philosophy of selection (non-agency), and the process of specialization (Polya-Hebbian dynamics). The fourth pillar completes the framework by providing the physical mechanism of selection: how does the environment select among units without a reward function or human evaluator? The answer draws on Karl Friston's Free Energy Principle (FEP) \cite{Friston2010,Friston2013}, which proposes that any self-organizing system maintaining its structural integrity over time does so by minimizing variational free energy, a quantity measuring the divergence between the system's internal model of the world and the actual sensory evidence it receives. Systems that successfully predict their sensory inputs persist; systems that fail to predict them experience escalating surprise (in the information-theoretic sense), which degrades their structural coherence until they dissolve. 

The critical insight for RECLAIM is that the Free Energy Principle is not a goal or reward but a physical law, or more precisely, a mathematical consequence of what it means for a bounded system to maintain a distinct identity in a changing environment. Friston (2013) \cite{Friston2013} argues that any system possessing a Markov blanket will, under certain regularity conditions, behave as if it is minimizing free energy or it will cease to exist as a distinct entity. While this claim has generated substantive debate regarding its scope and mathematical assumptions \cite{Biehl2021}, RECLAIM sidesteps this controversy entirely: the energy equation presented in Section~5.3.1 is a designed computational physics, stipulated by the researcher, not derived from first principles as a universal law of nature. Clark (2013) has explored the implications of this framework for biological cognition, characterizing the brain as fundamentally a prediction machine that continuously generates predictions about incoming sensory data and updates its internal model when those predictions fail \cite{Clark2013}.

In the existing AI literature, the Free Energy Principle has been applied primarily as a framework for designing individual agents. Work on active inference \cite{Friston2017,DaCosta2020} uses FEP to derive policies for agents acting in Markov decision processes, replacing reward maximization with free energy minimization as the agent's objective. While this represents a meaningful departure from standard reinforcement learning, it retains the fundamental structure of the engineering paradigm: a single agent optimizing a single objective. 

RECLAIM uses the Free Energy Principle in a fundamentally different way. Rather than implementing FEP as the objective function of an individual agent, RECLAIM implements FEP as the physics of the computational environment. Processing data costs computational energy; efficient processing (high compression, low prediction error) generates an energy surplus; inefficient processing generates an energy deficit; and units whose energy balance reaches zero dissolve. The researcher does not tell units to minimize free energy. Units that happen to minimize free energy persist, and those that do not cease to exist.

\subsection{The Unified Framework}

The four pillars form a single, integrated framework in which each pillar depends on and reinforces the others. BVSR provides the variation mechanism through random mutations that produce a diverse population of configurations. The Free Energy Principle provides the selection mechanism through which units that efficiently process information persist while those that fail dissolve. Polya-Hebbian dynamics provide the retention mechanism through which successful processing pathways are reinforced and locked in. And non-agency provides the philosophical constraint binding the other three together: the variation is blind, the selection is environmental, and the retention is path-dependent. The result is a complete alternative to the engineering paradigm, one in which gradient descent is replaced by blind variation, RLHF is replaced by environmental physics, identical static copies are replaced by unique path-dependent individuals, and teleological design is replaced by non-agentic emergence.

\section{Related Work}

The ideas underlying RECLAIM build upon decades of research in evolutionary computation, artificial life, neuroevolution, open-ended search, and active inference. In this section, we survey the most relevant prior systems, identifying what RECLAIM inherits from each and where it departs from or extends existing work.

\subsection{Digital Evolution and Neuroevolution}

The foundational work in computational ecology was Thomas Ray's Tierra system \cite{Ray1991}, in which self-replicating programs competed for CPU time and memory within a shared computational environment. Tierra demonstrated that digital organisms could spontaneously evolve complex ecological dynamics including parasitism and hyperparasitism, establishing the critical precedent that evolutionary dynamics are substrate-neutral. Avida \cite{Ofria2004} refined this approach by introducing isolated organisms with defined input-output behavior and a richer set of computational tasks, becoming the premier experimental platform for studying digital evolution. 

RECLAIM inherits from both systems the core concept of self-replicating digital organisms competing within a resource-constrained computational environment but departs in three significant ways: RECLAIM operates on neural network units with continuous-valued parameters and evolvable topologies rather than assembly-language programs, it eliminates the predefined evaluation tasks that Avida reintroduces, and it adds the Polya-Hebbian dynamics and cognitive food chain mechanisms absent from both systems.

In the domain of neuroevolution, the NeuroEvolution of Augmenting Topologies (NEAT) algorithm \cite{Stanley2002} introduced a principled approach to evolving both the weights and topology of neural networks, beginning with minimal networks and incrementally adding structural complexity through mutation. Subsequent work by Such et al. \cite{Such2017} and Salimans et al. \cite{Salimans2017} demonstrated that evolutionary strategies applied to networks with millions of parameters could achieve competitive performance on reinforcement learning benchmarks, challenging the assumption that gradient-based methods are strictly necessary for training large neural networks. 

RECLAIM draws directly on NEAT's topology evolution operators but departs from neuroevolution in a fundamental architectural respect: standard neuroevolution evaluates each network against a fitness function, while RECLAIM replaces explicit fitness evaluation with environmental energy economics, so that selection emerges as a consequence of thermodynamic viability rather than a human-specified performance metric. 

Karl Sims \cite{Sims1994} demonstrated the power of embodied evolution in his seminal work on Evolved Virtual Creatures, showing that populations of virtual organisms placed in simulated physical environments could spontaneously evolve complex morphologies and locomotion strategies without top-down design. RECLAIM conceptually extends Sims's achievement from the physical to the cognitive domain: where Sims evolved physical structures to navigate physical physics, RECLAIM evolves cognitive structures to navigate information physics, with the underlying principle remaining identical.

\subsection{Open-Ended and Co-Evolutionary Search}

Perhaps the closest philosophical ancestor of RECLAIM's non-agentic approach is novelty search \cite{Stanley2011}, an evolutionary algorithm that explicitly abandons objective functions in favor of selection for behavioral novelty. In deceptive optimization landscapes, novelty search consistently outperformed objective-driven search. The quality-diversity family of algorithms, exemplified by MAP-Elites \cite{Mouret2015}, extends this approach by seeking to fill a behavioral space with the highest-performing solution at each point. 

RECLAIM shares with these approaches the commitment to abandoning predefined objectives but goes further: novelty search replaces one objective (task performance) with another (behavioral novelty), while RECLAIM eliminates optimization metrics entirely, with novelty arising as a by-product of blind variation and co-evolutionary pressure rather than as an explicit selection criterion.

The Paired Open-Ended Trailblazer system (POET; \cite{Wang2019}) represents the most direct precedent for RECLAIM's co-evolutionary approach, simultaneously evolving both agents and their environments. RECLAIM extends POET in several directions: the distinction between agent and environment is blurred in RECLAIM because every unit's outputs become part of the environment for other units, RECLAIM eliminates fitness evaluation entirely, and RECLAIM's Polya-Hebbian dynamics introduce a developmental dimension absent from POET's purely evolutionary framework. 

Lenia \cite{Chan2019}, a continuous generalization of cellular automata producing remarkably lifelike self-organizing patterns, demonstrates that complex emergent behavior can arise from simple local rules in continuous substrates, supporting the plausibility of RECLAIM's emergence-based approach. However, Lenia focuses primarily on morphological complexity rather than information-processing complexity; RECLAIM addresses this gap by embedding neural processing within each unit and providing data streams that specifically reward information-processing capability. More recently, Kumar et al. \cite{Kumar2024} introduced ASAL (Automated Search for Artificial Life), a framework that uses vision-language foundation models to automate the discovery of novel structures in ALife substrates including Lenia, Boids, and neural cellular automata. ASAL demonstrates that automated search can discover lifeforms and emergent behaviors beyond those found through manual exploration, reinforcing the viability of open-ended computational ecologies. RECLAIM differs from ASAL in that it does not use an external foundation model as an evaluator; selection in RECLAIM is performed entirely by the internal energy economics, preserving the non-agentic constraint.

The broader question of open-endedness has become a major research focus in recent years. Banzhaf et al. \cite{Banzhaf2016} surveyed the requirements for open-ended novelty generation, Stanley et al. \cite{Stanley2017} proposed open-endedness as a grand challenge for AI, and Clune \cite{Clune2019} proposed AI-Generating Algorithms as an alternate paradigm for producing general intelligence through the co-evolution of environments, learning algorithms, and neural architectures. RECLAIM addresses this challenge through two mechanisms: the cognitive food chain ensures that the fitness landscape is perpetually shifting through co-evolutionary pressure, and the Polya-Hebbian dynamics create path-dependent trajectories that amplify small differences into qualitatively different cognitive architectures.

\subsection{Active Inference, Basal Cognition, and Mortal Computation}

The application of Friston's Free Energy Principle to artificial intelligence has generated a growing body of work under the umbrella of active inference \cite{Friston2017, DaCosta2020, Heins2022}. While active inference provides an elegant alternative to reward-based reinforcement learning, existing applications retain the fundamental structure of the engineering paradigm: a single agent is designed to minimize free energy as its objective function.  

RECLAIM uses the Free Energy Principle in a fundamentally different way, implementing FEP as the physics of the computational environment rather than as the objective of individual agents. We note that Friston himself characterizes the FEP as a physical principle rather than an agent objective; the novelty we claim is therefore not conceptual but architectural. While Friston frames FEP as physics in the theoretical sense, existing AI applications of active inference nonetheless implement it as an explicit optimization target for individual agents. RECLAIM, by contrast, designs a computational environment whose physics instantiate FEP consequences, so that units must minimize free energy to survive without being instructed to do so. This architectural implementation appears, to our knowledge, to be novel in the literature, preserving the mathematical content of the free energy framework while eliminating the teleological structure that agent-level implementations retain.

Michael Levin's research on basal cognition demonstrates that intelligence exists at the cellular and morphological level, with tissues and cell clusters possessing competency architectures that solve complex problems without central neural coordination \cite{Levin2021}. RECLAIM's bottom-up architecture aligns with Levin's findings by embedding cognitive potential within minimal, locally interacting autopoietic units. 

Geoffrey Hinton's concept of mortal computation, elaborated alongside his Forward-Forward algorithm \cite{Hinton2022}, argues that to achieve biological levels of energy efficiency, AI must abandon the paradigm of running identical, immortal software models on generic hardware, instead embracing systems where learned knowledge is inextricably bound to its specific physical substrate. RECLAIM is a digital manifestation of mortal computation: because each unit's cognitive architecture is uniquely shaped by its path-dependent Polya-Hebbian history, a unit cannot simply be copied or downloaded into a fresh state without losing its functional identity.

\subsection{Summary: What RECLAIM Adds}

The following table summarizes the relationship between RECLAIM and the existing systems surveyed above.

\begin{table}[htbp]
\centering
\footnotesize
\begin{tabular}{lccccccc}
\toprule
\textbf{Feature} & Tierra/Avida & NEAT & Novelty Search & POET & Lenia & Active Inf. & \textbf{RECLAIM} \\
\midrule
Self-replicating organisms & Yes & No & No & No & Partial & No & \textbf{Yes} \\
No evaluative objective & No & No & Partial & No & N/A & No & \textbf{Yes} \\
Neural topology evolution & No & Yes & No & No & No & No & \textbf{Yes} \\
Co-evolution (agents \& env.) & Partial & No & No & Yes & No & No & \textbf{Yes} \\
Free Energy Principle & No & No & No & No & No & Yes  & \textbf{Yes} \\
Polya-Hebbian specialization & No & No & No & No & No & No & \textbf{Yes } \\
Cognitive food chains & No & No & No & No & No & No & \textbf{Yes} \\
Data-as-sunlight & Partial & No & No & No & No & No & \textbf{Yes} \\
\bottomrule
\end{tabular}
\caption{Comparison of RECLAIM with prior systems.}
\label{tab:related_work}
\end{table}

Four contributions appear, to our knowledge, to be novel to the RECLAIM framework: the application of Polya-Hebbian dynamics as the mechanism for emergent specialization and digital individuality, the use of the Free Energy Principle as environmental physics rather than as an agent objective, the cognitive food chain as a mechanism for open-ended co-evolutionary complexity growth, and the conceptualization of data streams as environmental energy rather than as training curriculum.

\section{The RECLAIM Architecture: A Blueprint for Machine Life}

Having established the theoretical foundations and positioned them within the existing literature, we now present the RECLAIM architecture in full. RECLAIM is not a model to be trained but an ecology to be cultivated. It consists of three elements: a population of minimal computational units, a set of environmental physics governing their interactions, and an evolutionary engine that generates variation and enables selection. None of these components specifies what the system should learn, what structures it should develop, or what behaviors it should exhibit. These properties emerge from the interaction between units and environment over evolutionary time.

\subsection{Design Philosophy: Farming, Not Engineering}

The RECLAIM framework inverts the conventional relationship between the AI researcher and the intelligent system. In the engineering paradigm, the researcher is a designer: specifying architectures, defining objectives, curating training data, tuning hyperparameters, and evaluating outputs against benchmarks. The system is a product manufactured according to specification.

In RECLAIM, the researcher is a cultivator. The researcher designs the environment: its physical laws, its resource constraints, its data ecology, and its initial conditions. The researcher then seeds a population of simple, identical computational units into this environment and allows evolutionary dynamics to unfold. The researcher does not specify what the units should learn, how they should organize, or what cognitive structures they should develop. The researcher specifies only the conditions under which learning, organization, and cognitive structure become necessary for survival.

This distinction parallels the difference between engineering a machine and cultivating a living system. An engineer who builds a jet engine specifies every component, every tolerance, and every material. A farmer who grows a vineyard controls the soil composition, the irrigation schedule, the sunlight exposure, and the variety of grape planted, but the vine grows itself. The resulting vineyard exhibits complexity, resilience, and adaptive responsiveness to its environment that no engineer designed, because these properties emerged from the interaction between the organism and the conditions in which it was cultivated.

The RECLAIM researcher controls the equivalent of soil, water, sunlight, and seed selection. The researcher controls what data enters the system, how much computational energy is available, what the energy costs of processing are, how rapidly mutations occur, and what the initial population looks like. The researcher does not control what specializations emerge, what internal architectures develop, what cognitive strategies prove successful, or what the system ultimately becomes. These outcomes are products of evolutionary history, and they will differ across independent runs of the same system, just as two forests planted with the same seeds in the same soil will nonetheless grow into distinct ecosystems shaped by the contingencies of weather, insect populations, and microbial activity.

\subsection{The Autopoietic Unit: The Minimal Element of Digital Life}

\begin{figure}[htbp]
    \centering
    \includegraphics[width=0.8\textwidth]{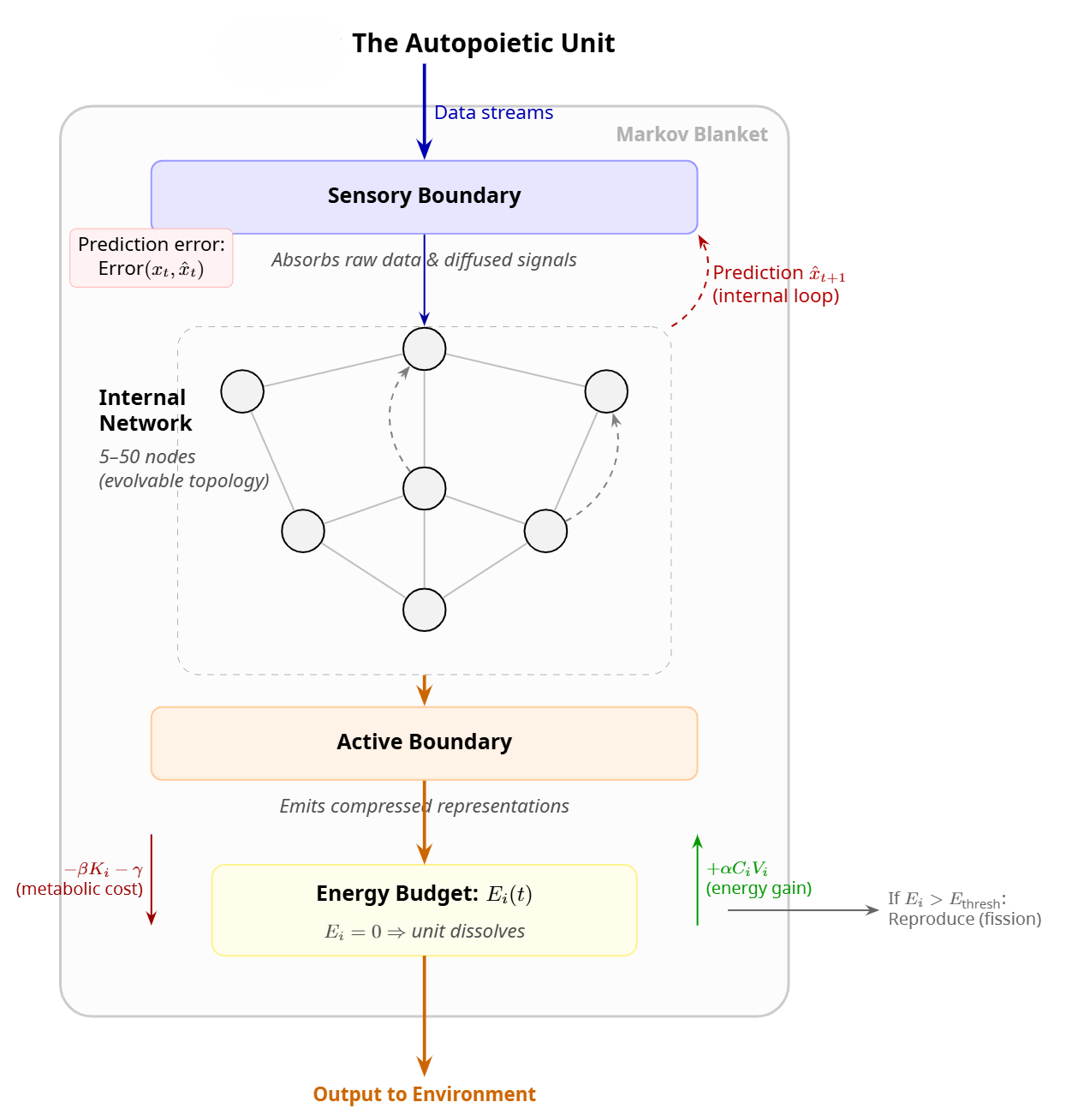}
    \caption{The Autopoietic Unit. A diagram illustrating a single unit with its sensory boundary absorbing data, the internal recurrent network processing the signals, the active boundary emitting compressed representations, and the energy pool regulating survival and reproduction.}
    \label{fig:autopoietic_unit}
\end{figure}

The fundamental building block of a RECLAIM ecology is the autopoietic unit. The term autopoietic, drawn from the work of Maturana and Varela \cite{Maturana1980}, denotes a system that is capable of maintaining and reproducing itself. Each unit is the simplest possible computational entity that satisfies this criterion: it can absorb data from its environment, process that data through an internal network, emit outputs, and maintain an energy budget that determines its continued existence. We adopt autopoiesis here in the operational sense discussed by McMullin (2004) \cite{McMullin2004}: the unit is autopoietic in that its continued existence depends on its own active information processing, not in the stronger claim that it physically produces its own computational substrate. Whether digital systems satisfy the original Maturana-Varela criteria, which require physical self-production of components, remains contested \cite{Bourgine2004}, and we use the term throughout to denote functional self-maintenance rather than to make ontological claims about digital life.
Each unit possesses three functional boundaries, corresponding to Friston's \cite{Friston2013} concept of the Markov blanket (see Figure \ref{fig:autopoietic_unit}).

The sensory boundary absorbs data from the external environment. This boundary determines what information the unit can perceive. It is the unit's interface with the incoming data streams that constitute the ecology's energy source.
The active boundary emits processed outputs back into the environment. These outputs become part of the environment for other units, creating the conditions for inter-unit interaction and, eventually, for the emergence of cognitive food chains (described in Section 5.6).

The internal network sits between the sensory and active boundaries. It consists of a small number of interconnected nodes (initially between 5 and 50) that process incoming data and generate outgoing signals. To enable temporal prediction of sequential data streams, the unit's internal network permits recurrent connections, allowing past activations to serve as inputs to current processing cycles. Topologies are essentially dynamic recurrent neural networks or discrete time continuous state spiking networks, giving them the requisite internal state to model temporal dependencies. The architecture of this internal network, including its weights, connectivity pattern, and number of nodes, constitutes the unit's genome and is subject to mutation during reproduction.

In addition to these three boundaries, each unit maintains an energy budget: a scalar quantity that increases when the unit efficiently processes data and decreases with every computational operation the unit performs. The energy budget is the mechanism by which the Free Energy Principle operates as environmental physics. When a unit's energy budget reaches zero, the unit is removed from the ecology. It has dissolved.

A crucial clarification regarding the internal processing architecture is required. Each unit's internal network produces two distinct outputs per processing cycle, serving fundamentally different functions. The first is an internal predictive reconstruction: the network generates a prediction of the next incoming data window, which is compared against the actual data that subsequently arrives at the sensory boundary to compute the self-prediction error (as formalized in Section 5.3.1). This prediction never leaves the unit; it is used solely for internal energy accounting. The second is an external emission: the network produces a compressed representation of its processed data, which is emitted through the active boundary into the environment and becomes available to neighboring units as a potential energy source. These two outputs may differ in dimensionality, format, and content. The predictive reconstruction must match the dimensionality of the incoming data (to permit error computation), while the external emission can take any format, with its dimensionality, value range, and temporal structure determined by evolvable boundary parameters subject to interface viscosity constraints (described in Section 5.6). This dual-output architecture ensures that the self-prediction error mechanism requires no external evaluator or top-down loss function: the unit's own future sensory experience serves as the sole reference signal.

Several properties of the unit design are critical to the RECLAIM framework:

\begin{itemize}
    \item \textbf{Units are autopoietic.} They must actively process data to maintain their energy budgets. A unit that ceases to process data will be consumed by maintenance costs and will dissolve. Existence requires activity.
    
    \item \textbf{Units are bounded.} The Markov blanket ensures that each unit's internal states are statistically separated from the external environment. The unit can only perceive the world through its sensory boundary and can only act on the world through its active boundary. This boundary is what makes the unit a distinct entity rather than merely a region of the computational substrate.
    
    \item \textbf{Units are simple.} A single unit with 5 to 50 internal nodes is not capable of sophisticated information processing. Complex cognition in RECLAIM does not arise from the complexity of individual units but from the interactions among many units within an ecology. This is analogous to biological neural networks: a single neuron is a simple threshold device, but networks of billions of neurons produce the full range of human cognitive capability.
    
    \item \textbf{The initial population is uniform.}  In the foundational generation, every unit begins with the same architecture: the same number of internal nodes, the same connectivity pattern, and randomly initialized weights drawn from the same distribution. Differentiation arises entirely through the Polya-Hebbian dynamics described in Section 3.3: random early processing successes are reinforced, driving each unit toward a unique, path-dependent specialization. No designer assigns roles or specializations. They emerge.
\end{itemize}

\subsection{The Digital Physics: The Economic Substrate}

The environment in which RECLAIM units exist is governed by a set of physical laws analogous to the thermodynamic constraints that govern biological life. These laws are not metaphors; they are the computational rules that are literally implemented in the simulation and that determine, with the same deterministic rigor as physical laws, what persists and what dissolves.

The first law is non-equilibrium energy flow. The RECLAIM ecology is an open, non-equilibrium thermodynamic system. Energy enters the ecology continuously via data streams (the digital equivalent of sunlight, described in Section 5.4) and exits the ecology continuously via metabolic maintenance costs (the digital equivalent of metabolic heat loss). The ecology is not a closed system with fixed total energy; it is a flow-through system in which energy enters, is processed, and dissipates. However, the computational substrate on which the ecology runs is finite and fixed: the total GPU memory, tensor capacity, and processing bandwidth available per timestep are bounded constants. Units compete not for shares of a fixed energy pool but for access to the finite data streams and the finite computational bandwidth of the substrate. This distinction between flowing energy and fixed substrate is critical: it ensures that the ecology operates far from thermodynamic equilibrium, which is precisely the condition under which self-organizing, dissipative structures \cite{Prigogine1977} can form and persist.

The second law is metabolic cost. Every computational operation that a unit performs, from absorbing data through its sensory boundary to propagating signals through its internal network to emitting outputs through its active boundary, costs energy. Furthermore, merely existing costs energy: each unit incurs a maintenance cost per timestep, representing the baseline computational overhead of maintaining its structural integrity. This maintenance cost ensures that passive existence is not a viable strategy; units must actively process data to offset their metabolic expenditure.

The third law is energy acquisition through information processing. Units gain energy by efficiently processing the data streams that flow through the ecology. The energy gained from processing a given data stream is proportional to the unit's compression efficiency: the ratio of the original data volume to the volume of the unit's internal representation, weighted by the accuracy of that representation. Crucially, data streams are subject to local depletion \cite{Tilman1982}. A specific spatial coordinate provides a finite maximum data volume $V_{\max}$ per timestep. If multiple units cluster in the same spatial radius to process the same easy data stream, the available energy is divided among the local population. This density-dependent resource scarcity prevents the entire population from stagnating on the simplest cognitive tasks. It creates the conditions for competitive exclusion, mathematically forcing mutant units to migrate and evolve the capacity to exploit harder, unoccupied data niches (such as text or temporal streams) where the energy-per-capita is higher.

The fourth law is death through energy depletion. When a unit's energy budget reaches zero, it is removed from the ecology, and its computational resources are freed for reallocation. This is the digital equivalent of biological death. When a unit dissolves, it leaves an empty spatial slot on the toroidal grid. No evaluator decides which units should die. Death is an automatic consequence of failing to maintain a positive energy balance under the constraints of metabolic cost and energy acquisition.

The fifth law is reproductive cost. A unit can reproduce (create a mutated copy of itself) only when its energy budget exceeds a specified threshold, typically set at twice the starting energy. Reproduction divides the parent's energy equally between parent and offspring. Units possessing an energy budget above the reproduction threshold may undergo fission, placing a mutated offspring into an adjacent empty slot. If no empty slots are adjacent, reproduction is deferred. This ensures that reproduction is costly, that only units with substantial energy surpluses can afford it, and that population density naturally regulates itself.

The sixth law is spatial locality. The RECLAIM ecology is not an undifferentiated global space. Units are embedded in a two-dimensional toroidal lattice, a discrete grid with periodic boundary conditions in which opposite edges connect, so that each unit occupies a specific lattice cell identified by integer coordinates $(x, y)$. This topology has two critical consequences. First, a unit's sensory boundary can absorb data only from sources within its local neighborhood, defined by a perception radius $r_s$ measured in lattice cells. Raw data streams are distributed across the surface as spatial gradients (analogous to weather systems moving across terrain), so that different regions of the ecology receive different data at different times. Second, a unit's active boundary outputs attenuate with distance: a unit's emitted signal is received by neighboring units at a strength inversely proportional to the squared distance between them, and signals beyond the attenuation radius $r_a$ are not received at all. 

This spatial locality constraint is not merely a computational optimization (though it reduces the interaction cost from $O(N^2)$ to $O(N \cdot k)$, where $k$ is the average neighborhood size). It is an ecological necessity. Without spatial structure, all units would receive identical global broadcasts, eliminating the local niche differentiation that drives speciation and trophic hierarchy. With spatial structure, different regions of the ecology can develop distinct local ecosystems, different data stream compositions can create different selection pressures at different locations, and units must physically migrate (through small random position perturbations at each timestep) to access different resources. The toroidal topology ensures that the surface has no edges, preventing boundary artifacts.

\subsubsection{Mathematical Formalization of the Digital Physics}

The six laws described above can be expressed in a compact mathematical form. Let \(E_i(t)\) denote the energy budget of unit \(i\) at timestep \(t\). The energy update at each timestep is governed by the following equation:

\begin{equation}
E_i(t+1) = E_i(t) + \alpha \, C_i(t) \, V_i(t) - \beta \, K_i(t) - \gamma
\end{equation}

where \(C_i(t)\) is the compression ratio achieved by unit \(i\) at timestep \(t\), defined as the ratio of input data volume to internal representation volume weighted by representational accuracy; \(V_i(t)\) is the volume of data processed; \(K_i(t)\) is the total computational cost of the processing cycle (proportional to the number of active connections and the number of internal propagation steps); and \(\gamma\) is the fixed maintenance cost. The constants \(\alpha\), \(\beta\), and \(\gamma\) are environmental parameters set by the researcher and held constant throughout the simulation.

A critical clarification is required regarding the term representational accuracy in the compression ratio \(C_i(t)\). A skeptical reader may object that computing accuracy against ground truth would reintroduce a top-down objective function, undermining the non-agentic claim. We therefore define accuracy strictly as an intrinsic, self-computed metric requiring no external evaluator. The prediction-error mechanism operates as a one-step-ahead prediction cycle with the following timing convention: at each timestep \(t\), the unit absorbs incoming data \(x_t\) and evaluates its previously generated prediction \(\hat{x}_t\) (produced at timestep \(t-1\)) against the actual arrival \(x_t\). This prediction error feeds into the compression ratio \(C_i(t)\), which in turn determines the energy update \(E_i(t+1)\). Simultaneously, the unit uses its current internal state to produce a new prediction \(\hat{x}_{t+1}\), which will be evaluated at the next timestep. For continuous streams (such as waveforms and numerical sequences), the prediction error is computed as the Mean Squared Error (MSE). For discrete streams (such as text and noise), the data is treated as raw binary bitstreams, and the error is computed using Binary Cross-Entropy (BCE) or Hamming distance.

This distinction is critical to prevent non-physical gradients; computing continuous MSE on discrete UTF-8 byte integers produces artificial distance metrics where predicting a lowercase letter instead of an uppercase letter is penalized far more severely than predicting an entirely unrelated character. No external ground truth is consulted, no human labels are referenced, and no teacher signal is provided. The environment simply delivers data, and the energy equation rewards units whose internal models produce predictions that closely match the next data arrival. This mechanism is structurally analogous to predictive coding \cite{Clark2013, Friston2010}: the unit gains energy in proportion to the accuracy of its own predictions about its own future sensory experience, computed entirely within the unit's own processing cycle.

The compression ratio \(C_i(t)\) can now be stated as a precise sub-equation:

\begin{equation}
C_i(t) = \left( \frac{V_{\text{in}}}{V_{\text{repr}}} \right) \exp\left( -\text{Error}(x_t, \hat{x}_t) \right)
\end{equation}

where \(V_{\text{in}}\) is the dimensionality of the incoming data window, \(V_{\text{repr}}\) is the number of active nodes in the unit's internal representation, and \(\text{Error}(x_t, \hat{x}_t)\) is the prediction error between the actual data \(x_t\) arriving at timestep \(t\) and the unit's prediction \(\hat{x}_t\) generated from its compressed representation at the previous timestep \(t-1\). A node \(j\) is classified as active at timestep \(t\) if its activation \(y_j(t)\) exceeds a minimum activity threshold \(\tau\), a fixed environmental parameter set by the researcher and not subject to evolutionary modification. Nodes with activations below \(\tau\) are classified as dormant and excluded from the \(V_{\text{repr}}\) count. If all nodes fall below the activity threshold, the unit has ceased active processing; it receives no energy gain (\(C_i(t) = 0\)) for that timestep, ensuring that the compression ratio remains well-defined and that complete network dormancy is metabolically equivalent to processing failure. This threshold prevents units from gaming the dimensionality ratio by maintaining nodes at negligible activation values.

We note that the ratio \(V_{\text{in}} / V_{\text{repr}}\) measures architectural parsimony rather than information-theoretic compression in the strict sense of Shannon (1948). An ideal compression ratio would require computing the mutual information between the input and the internal representation, a quantity that is computationally intractable for arbitrary networks. The dimensionality ratio serves as an efficiently computable proxy that rewards representational economy: units that achieve accurate prediction with fewer active nodes receive higher energy returns. While this proxy does not constitute a formal rate-distortion measure, it is consistent with the minimum description length principle \cite{Rissanen1978}, which equates model quality with the parsimony of the encoding.

A critical mathematical safeguard against trivial prediction loopholes exists within the broader energy equation. If a unit attempts to minimize prediction error simply by passing the input directly to the output without compression, the metabolic cost of maintaining the active internal nodes required for that pass-through processing will rapidly deplete its energy budget. Conversely, a unit might attempt to game the dimensional ratio by deleting its internal network entirely, reducing \(V_{\text{repr}}\) to 1. To prevent this trivial collapse, the environmental parameters \(\alpha\), \(\beta\), and \(\gamma\) must satisfy a formal constraint. Let \(\text{Error}_{\text{baseline}}\) denote the prediction error achieved by a naive strategy (pass-through or trivial network). Trivial collapse is prevented if and only if \(\alpha \, V_{\text{in}} \, \exp(-\text{Error}_{\text{baseline}}) < \gamma\), ensuring that the energy gained from a high dimensionality ratio under poor prediction is strictly less than the maintenance cost. Under this constraint, a high prediction error drives the \(\exp(-\text{Error})\) term effectively to zero, completely neutralizing the massive \((V_{\text{in}} / 1)\) multiplier. Energy is only awarded for actual compression of successfully predicted data.

A third potential gaming strategy warrants explicit consideration: narrow memorization. A unit might evolve to memorize short repeating patterns in a stationary data stream, achieving near-zero prediction error on those specific patterns with very few active nodes and thereby producing an inflated compression ratio. This would constitute the RECLAIM equivalent of overfitting. The ecology defends against this through two mechanisms. First, data streams are distributed as continuous spatial gradients that shift over time (analogous to weather systems; see Section~5.3), so that the statistical properties of the data arriving at any given spatial coordinate are non-stationary; a unit that memorizes a specific pattern will suffer escalating prediction errors as the local data distribution drifts. Second, the density-dependent resource depletion described in the third law ensures that if multiple memorization-specialized units cluster on the same predictable data stream, the available energy per unit declines below maintenance cost, selecting against narrow specialization on easy patterns.

An important question concerns the sensitivity of the ecology to these parameter choices. We argue that the qualitative behavior of the system is robust across a range of parameter settings, provided that a fundamental thermodynamic constraint is preserved: energy gain from genuine compression must exceed maintenance cost for well-adapted units, while energy gain from trivial strategies (pass-through or network collapse) must fall below maintenance cost. Within this constraint envelope, the specific values of \(\alpha\), \(\beta\), and \(\gamma\) will influence the pace of evolution and the stringency of selection but will not determine the qualitative character of the emergent dynamics, analogous to how varying gravitational strength across planets changes the specific morphologies that evolve but does not change the fundamental logic of natural selection. Characterizing the precise boundaries of this constraint envelope through systematic parameter sweeps is a central objective of the planned experimental program.

The activity threshold \(\tau\) is fixed as an environmental constant rather than subject to evolutionary modification for the same reason that gravitational acceleration is not subject to biological evolution: it is part of the physics within which evolution operates, not part of the phenotype that evolution shapes. Allowing \(\tau\) to evolve would permit units to trivially inflate their compression ratio by redefining what counts as an active node, collapsing the energy equation's capacity to distinguish genuine compression from definitional manipulation.

This formulation bears a formally precise relationship to variational inference that extends beyond qualitative analogy. In the ELBO framework, model quality is assessed by balancing reconstruction accuracy (expected log-likelihood) against model complexity (KL divergence from the prior). The connection to Equation~2 can be stated exactly: when the prediction error is Mean Squared Error (for continuous streams), \(\exp(-\text{MSE})\) is proportional to the Gaussian likelihood \(p(x_t | \hat{x}_t)\) under a unit-variance assumption; when the error is Binary Cross-Entropy (for discrete streams), \(\exp(-\text{BCE})\) equals the Bernoulli likelihood \(p(x_t | \hat{x}_t)\) exactly. The compression ratio therefore has the structure of Bayesian model evidence: the dimensionality ratio \((V_{\text{in}} / V_{\text{repr}})\) serves as an Occam factor penalizing model complexity (analogous to the prior's preference for simpler models), while \(\exp(-\text{Error})\) serves as the data likelihood. The product \(C_i(t) = \text{Occam Factor} \times \text{Likelihood}\) thus mirrors the exact form of marginal likelihood in Bayesian model selection. We note that this correspondence is exact for the likelihood term but approximate for the complexity term, since the true Bayesian Occam factor depends on the full parameter-space geometry rather than on a simple node count. Nevertheless, the structural parallel is sufficiently precise that energy gain in the RECLAIM ecology can be understood as a computationally tractable approximation to Bayesian model evidence, with units that maximize model evidence persisting and those that do not dissolving.

The connection between this energy equation and the Free Energy Principle can be stated precisely. In Friston's formulation, variational free energy \(F\) is defined as the divergence between the system's internal model \(Q\) and the true posterior distribution \(P\) over environmental states, given sensory observations. Minimizing \(F\) is mathematically equivalent to maximizing the evidence lower bound (ELBO). In the RECLAIM energy equation, the compression ratio \(C_i(t)\) approximates this quantity. Thus, the energy gain term \(\alpha \, C_i(t) \, V_i(t)\) is a monotonically decreasing function of variational free energy. Units that minimize free energy maximize energy gain. Units that fail to minimize free energy deplete their budgets and dissolve. The FEP is not a goal the units pursue; it is a mathematical consequence of the energy equation that governs their existence.

A legitimate objection arises at this point: does the energy equation not constitute a fitness function by another name? We acknowledge that the energy equation defines the selection criterion, and in this formal sense it plays the role that a fitness function plays in conventional evolutionary computation. The distinction we draw is not that RECLAIM lacks a selection criterion, for all evolutionary systems require one, but that the selection criterion is environmental rather than evaluative. In conventional evolutionary computation, a fitness function computes how well an organism achieves a human-specified goal and assigns a corresponding score. In RECLAIM, the energy equation computes whether an organism can sustain itself within the physics of its environment, with no reference to any human-specified goal. The difference is analogous to the distinction between a teacher grading an examination (evaluative selection against a human-defined rubric) and gravity acting on physical objects (environmental selection with no reference to any purpose). Both select, but only the former involves a proxy for human intent, and only the former is subject to Goodhart's Law. Goodhart's Law operates specifically when there exists a gap between what is measured (the proxy) and what is intended (the true goal). In RLHF, the reward model score is a proxy for human satisfaction, and the gap between them is precisely what enables reward hacking. In RECLAIM, the energy equation measures thermodynamic viability, and there is no separate intended outcome for it to diverge from; the equation is not a proxy for something else but the complete specification of the physics. A unit cannot exploit a proxy-intent gap in the energy equation, because the energy equation contains no reference to human intent and therefore presents no such gap. We note, however, that while this eliminates the specific proxy-intent divergence that enables reward hacking in RLHF systems, it does not guarantee that thermodynamic viability will select for cognitive complexity rather than minimal survival strategies such as memorization of short-range patterns or metabolic stasis on easy data streams. The cognitive food chain described in Section~5.6, with its co-evolutionary Red Queen dynamics, provides the hypothesized mechanism for driving complexity beyond metabolic minimalism, but whether this suffices to produce hierarchical cognitive architecture remains an empirical question that the planned experimental program is designed to address. The energy equation is a designed physics, and the choice of its parameters constitutes a consequential design decision that constrains the space of possible ecologies.

The Polya-Hebbian update rule can similarly be formalized. Let \(w_{ij}\) denote the connection weight from node \(j\) to node \(i\) within a given unit's internal network. After each processing cycle, the weight is updated according to:

\begin{equation}
\Delta w_{ij} = \eta \, S_i(t) \, x_j \, y_i - \lambda \, w_{ij}
\end{equation}

where \(\eta\) is the Hebbian learning rate, \(S_i(t)\) is the energy surplus (or deficit) generated by the processing cycle (\(S_i(t) = \alpha \, C_i(t) \, V_i(t) - \beta \, K_i(t)\)), intentionally omitting the static maintenance cost \(\gamma\) because including it would cause all weight updates to be negatively biased regardless of processing quality, systematically weakening all pathways including successful ones; \(x_j\) is the activation of the presynaptic node \(j\) and \(y_i\) is the activation of the postsynaptic node \(i\), both computed through a rectified linear activation function that guarantees non-negative values and ensures that the sign of the product \(S_i(t) \cdot x_j \cdot y_i\) is determined solely by the energy surplus term; and \(\lambda\) is a small weight decay constant that prevents unbounded weight growth, analogous to the normalization mechanisms explored by Oja \cite{Oja1982}. This non-negativity constraint is essential to the mixed-sign reinforcement dynamics described below: without it, negative activations could invert the intended direction of synaptic change, producing reinforcement when weakening was intended. We note that while activations are constrained to be non-negative through ReLU, the connection weights \(w_{ij}\) themselves are unconstrained real-valued parameters that may be positive (excitatory) or negative (inhibitory). When \(S_i(t) < 0\), the update rule weakens co-active connections, potentially driving their weights through zero to become inhibitory. The weight decay term \(-\lambda \, w_{ij}\) provides a natural restoring force toward zero for weights of either sign, preventing runaway excitation or inhibition. In practice, sustained negative energy surplus causes unit dissolution before weights can diverge, so that persistent negative weights arise only in units that alternate between surplus and deficit across processing cycles, producing the mixed excitatory-inhibitory connectivity observed in biological neural circuits.

This update rule constitutes a \emph{three-factor learning rule} in the terminology of modern computational neuroscience \cite{Fremaux2016}: the three factors are presynaptic activity \(x_j\), postsynaptic activity \(y_i\), and a global modulatory signal \(S_i(t)\) that gates the direction and magnitude of plasticity. It is formally equivalent to Hebb's principle \cite{Hebb1949}, with the crucial addition of the energy surplus as the third modulatory factor, analogous to the role of neuromodulators such as dopamine in biological synaptic plasticity. The decay term prevents runaway weight growth. We note that because \(S_i(t)\) can be negative, this constitutes a generalized reinforcement scheme in which the direction of synaptic change is modulated by thermodynamic success. When \(S_i(t) > 0\), the update is positively reinforcing (P\'{o}lya-like), strengthening active pathways and creating path-dependent lock-in. When \(S_i(t) < 0\), the update is negatively reinforcing, weakening active pathways and providing a self-corrective mechanism that prunes failed processing hypotheses. This mixed-sign reinforcement differs from both the classical P\'{o}lya urn (which permits only positive reinforcement) and the Friedman urn \cite{Friedman1949} (which is purely self-correcting and converges to deterministic equilibria). The path-dependent convergence properties of the classical model are preserved in the dominant regime because units that consistently generate energy deficits dissolve before negative reinforcement can drive their internal states to equilibrium; only units in the positive-reinforcement regime persist long enough for their specializations to lock in through the self-reinforcing dynamics described in Section~3.3.

\subsection{Data Streams as Sunlight: The Primordial Energy Source}

In Earth's biosphere, nearly all energy ultimately derives from sunlight. Photosynthetic organisms capture solar radiation and convert it into chemical energy, which then flows through the food chain to support all other forms of life. The sun does not instruct organisms on how to use its energy. It simply radiates, and the organisms that evolve the capacity to capture and convert that radiation persist.

\begin{figure}[htbp]
    \centering
    \includegraphics[width=0.7\textwidth]{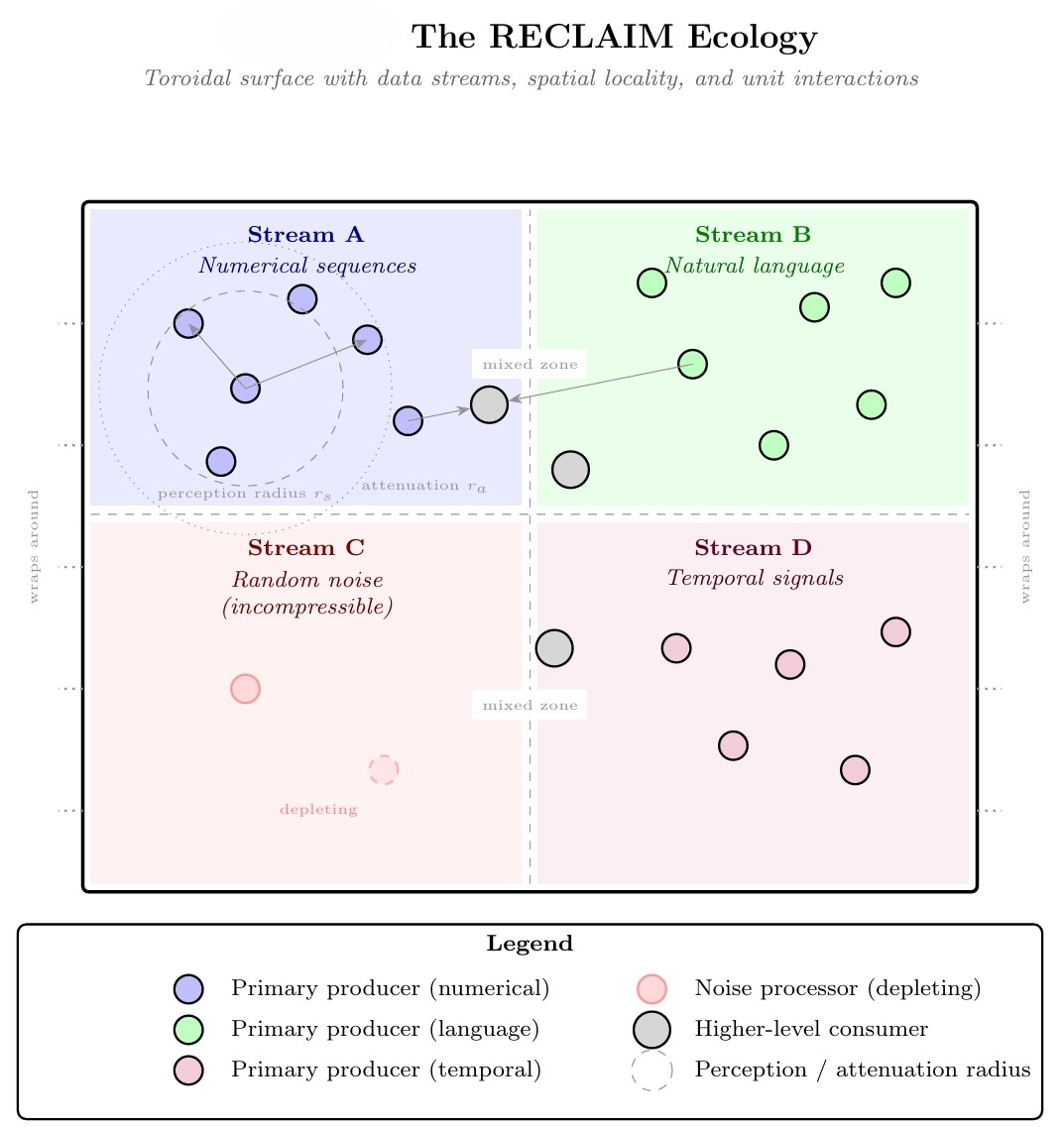}
    \caption{The RECLAIM Ecology. A visualization of the toroidal lattice, depicting multiple data streams distributed across the space, with individual units and spatial diffusion mediating interactions. For visual clarity, streams are shown as discrete regions; in implementation, stream intensities vary as spatial gradients analogous to weather systems, with overlapping transition zones between dominant stream types.}
    \label{fig:reclaim_ecology}
\end{figure}

In a RECLAIM ecology, data streams serve the equivalent function. Multiple data streams flow continuously into the ecology, each representing a different type of structured information (see Figure \ref{fig:reclaim_ecology}). In the proposed experimental configuration, four simultaneous streams would flow continuously into the ecology: structured numerical sequences (such as Fibonacci numbers and prime sequences), natural language text (drawn from curated corpora such as Wikipedia), random noise (Kolmogorov-random bitstrings that are, by mathematical definition, incompressible), and temporal signals (such as sinusoidal waveforms with varying frequency and amplitude).

These streams differ in their entropy and therefore in their energy potential. The numerical sequences have low entropy and high compressibility, making them rich potential energy sources for units that evolve the capacity to detect and exploit their statistical structure. The natural language text has moderate entropy, offering substantial but more challenging compression opportunities. The random noise has maximal entropy and is formally incompressible: no unit can extract sustained energy from it because it contains no exploitable structure. While finite segments of random noise may exhibit short-range statistical fluctuations that a unit could temporarily exploit through overfitting, these apparent patterns do not persist across timesteps, and a unit that specializes in noise processing will incur metabolic costs that exceed its brief energy gains, producing a net negative energy balance over any sustained period. Units that allocate processing resources to the noise stream will incur computational costs without any energy return, and they will eventually be outcompeted by units that learn to avoid it. The temporal signals offer energy to units that evolve the capacity for temporal prediction, rewarding a qualitatively different type of information processing than the static streams.

The specific composition and configuration of data streams is a primary design parameter controlled by the researcher, and different stream compositions will produce ecologies with qualitatively different emergent capabilities, just as different soil compositions and climates produce forests with different species profiles. The four streams proposed here are intended as a minimal experimental configuration sufficient to test the framework's core predictions, and the question of what stream compositions produce specific types of cognitive capability remains an empirical question for the experimental program.

A critical point of methodological discipline must be emphasized regarding the sensory interface. All data streams are presented to units as raw numerical arrays: normalized continuous-valued vectors derived directly from the underlying data without any human-engineered preprocessing. Numerical sequences are presented as raw floating-point values. Text is presented as raw byte sequences (UTF-8 encoded), not as tokenized words or pre-computed embeddings. Random noise is presented as raw bitstrings. Temporal signals are presented as raw amplitude values sampled at a fixed rate. No tokenization, no word embeddings, no feature extraction, and no dimensionality reduction are applied before the data reaches the units' sensory boundaries. This is a non-negotiable design constraint. If we were to feed units pre-tokenized representations (such as Word2Vec or transformer embeddings), we would be smuggling the very top-down engineering that RECLAIM is designed to circumvent into the system through its input layer. By forcing units to build their own representations directly from raw byte arrays, the framework addresses what we term the \emph{Tokenization Prior Problem}: the dependence of current AI systems on human-engineered tokenization schemes, embeddings, and feature extractors that smuggle implicit semantic commitments into the system before learning begins. While this does not constitute a solution to the full Symbol Grounding Problem as defined by Harnad \cite{Harnad1990}, which requires grounding in physical sensorimotor interaction with the world, it does ensure that RECLAIM units must evolve their own sensory parsing protocols, internal representations, and tokenization strategies from raw signals rather than inheriting human-engineered semantic priors. This is computationally demanding, but it is the only approach consistent with the non-agentic, bottom-up philosophy of the framework.

A critical distinction must also be drawn between data-as-environment and data-as-curriculum. In the current AI paradigm, data is curriculum: the model is explicitly instructed (through the objective function) to predict, classify, or generate from the data. The data defines the task, and the model is evaluated on its performance at that task. In RECLAIM, data is environment. The data flows through the ecology like sunlight through a forest. No unit is told what to do with the data. No unit is evaluated on how well it processes the data. Units that happen to evolve the capacity to extract energy from the data persist. Units that do not, dissolve. The distinction is the difference between a student memorizing a textbook because a teacher assigned it and a retina evolving the capacity to process photons because processing photons conferred a survival advantage. In the first case, the curriculum defines what is learned. In the second case, the environment creates the conditions under which learning becomes necessary for survival, but the specific form that learning takes is determined by evolutionary dynamics, not by curricular specification.

\subsection{The Evolutionary Engine: Variation, Selection, and Retention}

The evolutionary dynamics of a RECLAIM ecology implement the BVSR framework described in Section 3.1 through three concurrent processes: variation through mutation, selection through energy economics, and retention through Polya-Hebbian reinforcement.

\subsubsection{Variation through Mutation}

When a unit reproduces, its offspring inherits a mutated copy of the parent's genome. Four types of mutation operate simultaneously:

Weight mutation is the most common, occurring with a probability of approximately 0.01 per connection per reproduction event. Each connection weight in the offspring's internal network is randomly perturbed by a small amount drawn from a zero-mean Gaussian distribution. This introduces fine-grained variation into existing network structures.

Topological mutation occurs with lower probability (approximately 0.05 per reproduction event) but introduces qualitative structural changes. A topological mutation may add a new node to the network (by splitting an existing connection and inserting an intermediate node), add a new connection between previously unconnected nodes, or delete an existing connection. Over many generations, topological mutations enable the evolution of progressively more complex internal architectures, a mechanism inspired by the NEAT algorithm \cite{Stanley2002}.

Parameter mutation alters the unit's operational parameters, including its sensory sensitivity thresholds (which data streams it absorbs and at what rate), its processing speed and accuracy tradeoff, and its communication range (how far its outputs propagate through the ecology). These mutations affect the unit's ecological niche rather than its internal processing capacity.

Recombination occurs when two units with surplus energy reproduce together. The offspring's genome is constructed through random crossover of the two parent genomes, combined with independent mutations. Recombination introduces a qualitatively different source of variation from mutation alone, enabling the combination of beneficial traits that evolved independently in different lineages.

All four mutation types are blind. No mutation is directed toward any goal, and there is no mechanism for the system to preferentially generate mutations that are likely to be beneficial. This is the defining characteristic of BVSR: variation is genuinely random, and directionality emerges only through the cumulative effect of selection across many generations.

\subsubsection{Selection through Energy Economics}

Selection in a RECLAIM ecology is not an explicit process performed by an evaluator. It is an automatic consequence of the energy economics described in Section 5.3. At each timestep, every unit in the ecology executes a processing cycle:

The unit absorbs data from the environment through its sensory boundary. Its internal network processes the absorbed data, generating an internal representation and a prediction of future data. The unit emits its processed output through its active boundary. The energy budget is then updated: the unit gains energy proportional to its compression efficiency (the ratio of input data volume to internal representation volume, weighted by representational accuracy) and loses energy proportional to the computational cost of processing plus the fixed maintenance cost. If the unit's energy budget falls to zero, the unit is removed from the ecology.

This process requires no fitness function, no reward model, and no human evaluation. The environment itself performs selection through the automatic consequences of its physical laws. Units that compress data efficiently gain more energy than they spend, accumulate surplus energy, and eventually reproduce. Units that compress inefficiently, or that waste computational resources on incompressible noise, spend more energy than they gain, deplete their budgets, and dissolve. Over many generations, the population shifts toward configurations that are progressively better adapted to the information structure of the data ecology.

\subsubsection{Retention through Polya-Hebbian Reinforcement}

Within each unit, the Polya-Hebbian dynamics described in Section 3.3 operate at the level of individual processing cycles. After each cycle, the internal pathways that contributed to successful data compression (energy gain) are strengthened: their connection weights are multiplied by a factor slightly greater than one. Pathways that did not contribute to successful compression are weakened: their connection weights are multiplied by a factor slightly less than one, representing natural decay of unused connections.

This Hebbian reinforcement creates a positive feedback loop within each unit. Processing pathways that happen to be effective for a particular type of data become stronger, which makes the unit more likely to process that type of data successfully in the future, which reinforces those pathways further. Over time, this dynamic drives each unit toward specialization in a particular data type or processing function, even though no designer specified what that specialization should be. The specialization is an emergent consequence of Polya-Hebbian dynamics operating on the random initial conditions of each unit's processing history.

\subsubsection{The Baldwin Effect and Genetic Assimilation}

The interaction between within-lifetime Polya-Hebbian learning and across-generation BVSR mutation creates a powerful evolutionary accelerator known as the Baldwin Effect \cite{Baldwin1896}, demonstrated computationally by Hinton and Nowlan \cite{Hinton1987} in their landmark study showing that learning can smooth an otherwise intractable fitness landscape. The Polya-Hebbian bridge does not merely create individual specialization; it reshapes the evolutionary landscape for the entire population. When units consistently expend metabolic energy during their lifetimes to learn a specific processing trait via Hebbian reinforcement, a selective pressure is generated. Any random mutation that genetically hardwires that trait at birth will be strongly favored by environmental selection, because the mutated unit saves the metabolic cost of learning the trait and can dedicate that energy to further processing or reproduction. Over many generations, behaviors that are initially acquired through lifetime learning become genetically assimilated into the innate architecture of the population. This demonstrates that RECLAIM captures the full depth of evolutionary dynamics, where learning and evolution are not separate processes but deeply coupled mechanisms driving swift adaptation.

\subsection{The Cognitive Food Chain: The Red Queen Engine for Open-Ended Evolution}

Perhaps the most novel architectural feature of RECLAIM is the cognitive food chain, a mechanism designed to produce open-ended evolutionary complexity without requiring external environmental changes or predefined complexity ladders.

In the architecture described thus far, units absorb energy from raw data streams. However, RECLAIM units do not only consume raw data. They can also consume each other's outputs. When a unit processes data and emits its output through its active boundary, that output becomes part of the environment for nearby units (subject to the spatial attenuation defined in the sixth law of Section 5.3). If a second unit possesses a sensory boundary tuned to receive the first unit's output format, the second unit can absorb and process this output, potentially extracting energy from it if the output contains exploitable statistical structure.

A legitimate question arises regarding the combinatorics of the communication interface between units. If Unit A compresses data and emits a representation of a particular size and format through a subset of its active boundary, how does Unit B read that exact output without relying on astronomically improbable random mutations to discover the correct receptor mask? This problem is solved through a mechanism we term Spatial Chemical Diffusion or Signal Broadcasting. Instead of requiring Unit B to mutate the exact receptor mask matching Unit A's output, the unmasked signals from Unit A physically diffuse into the local environment grid. Unit B's receptors simply process whatever overlapping signals hit its boundary from the surrounding medium. When Unit B encounters these diffused signals, the standard Hebbian reinforcement dynamics naturally strengthen the internal pathways connected to whichever receptors happen to be receiving structured information. The combinatorial search problem is thus replaced by physical proximity and associative learning. This creates a robust sender and receiver co-evolutionary dynamic. Producers evolve output formats that maximize their own compression efficiency, and the diffusing signals automatically shape the receptor sensitivity of nearby consumers through Hebbian plasticity. The result is an emergent, co-evolved communication protocol that no designer specified. This adds a layer of complexity that enriches the ecology. Units do not only evolve to compress data, they must also evolve compatible communication channels to pass energy between trophic levels.

A related concern involves the stability of these evolved interfaces. If the dimensional structure of a producer unit's output changes too rapidly through mutation, consumer units will suffer catastrophic prediction errors, lose energy, and die before they can adapt. Biological systems face an analogous problem and solve it through slow-changing, universal chemical interfaces: ATP serves as the universal energy currency regardless of what organism produces or consumes it, and neurotransmitter types change on evolutionary timescales, not on individual-lifetime timescales. RECLAIM incorporates an analogous mechanism through what we term \emph{interface viscosity}. While internal network weights mutate at the standard rate defined in Section 5.5.1, the parameters that define the active boundary output format (its dimensionality, value range, and temporal structure) and the sensory boundary receptor format mutate at a substantially reduced rate, approximately one hundredth of the internal weight mutation rate. This differential mutation rate creates a stable interface layer that changes slowly relative to the internal processing architecture, giving consumer units evolutionary time to adapt to gradual changes in producer output formats while permitting swift internal optimization. The interface viscosity is analogous to the stability of physical constants in biological ecosystems: the laws of chemistry do not change on the timescale of individual adaptation, even though the organisms subject to those laws change rapidly.

Interface viscosity addresses the stability of output format (dimensionality, value range), but a subtler challenge involves semantic stability. Even if a producer unit's output format remains constant (for example, a 10-dimensional vector), a mutation in its internal network might alter the statistical meaning of the values within that vector. If a particular output pattern previously signaled the presence of periodic structure and, after mutation, signals something entirely different, consumer units that depend on the semantic content of that output will suffer unexpected prediction errors. We observe that the Red Queen dynamic described below provides an intrinsic pressure toward semantic smooth gradients. A producer unit whose internal mutations drastically alter the semantic content of its outputs will disrupt the consumer units that feed on those outputs, potentially causing local ecosystem collapse. The mechanism by which this ecosystem collapse affects the producer requires explicit clarification regarding the energy economics of the cognitive food chain.

A critical point of thermodynamic bookkeeping must be stated precisely. Energy in the RECLAIM ecology is \emph{not} conserved across trophic levels. When a Level~2 consumer processes the output of a Level~1 producer, it does not subtract energy from the producer. Rather, the consumer earns energy through the same compression-ratio mechanism (Equation~2) applied to the producer's output as a data source. The producer's output serves as an alternative data stream, and the consumer's energy gain is determined by how efficiently it compresses that stream. This mirrors biological thermodynamics: a herbivore does not remove the sun's energy from a plant; it captures free energy from the chemical bonds in the plant's biomass, while the plant continues to photosynthesize independently. However, the producer-consumer coupling is not energetically neutral for the producer. Consumer units that successfully process a producer's outputs occupy spatial slots in the producer's local neighborhood. Because spatial slots are finite and reproduction requires adjacent empty slots (Section~5.3), a thriving consumer population around a producer reduces competition from rival producer lineages that would otherwise occupy those slots, creates a richer local data environment through the consumers' own emissions, and generates higher-order ecological niches from which the producer's descendants may benefit. Conversely, the collapse of the local consumer population opens spatial slots to competing producers, increases local competition for raw data streams, and impoverishes the local data ecology. The selective pressure on producers to maintain semantic stability in their emissions therefore operates through ecosystem-mediated spatial competition rather than through direct energy transfer, analogous to how plants benefit from the pollinator populations they support without receiving direct energy from them.

There is therefore evolutionary pressure for internal mutations to improve processing efficiency while preserving the statistical regularities of external emissions. Mutations that improve compression without disrupting downstream consumers will be selected for, while mutations that cause semantic discontinuities will be selected against through the ecosystem-mediated consequences of consumer collapse. This creates a natural pressure for incremental, smooth semantic change rather than abrupt semantic shifts, analogous to the evolutionary stability of biochemical signaling pathways in biological ecosystems.

This creates the conditions for an emergent trophic hierarchy, analogous to the food chains observed in biological ecosystems. In a biological ecosystem, primary producers (such as plants) convert raw environmental energy (sunlight) into chemical energy through photosynthesis. Primary consumers (herbivores) feed on the producers, extracting energy from their biomass. Secondary consumers (predators) feed on the primary consumers, and tertiary consumers feed on the secondary consumers. Each trophic level processes the output of the level below it, creating a hierarchy of increasing abstraction and organizational complexity.

\begin{figure}[htbp]
    \centering
    \includegraphics[width=0.7\textwidth]{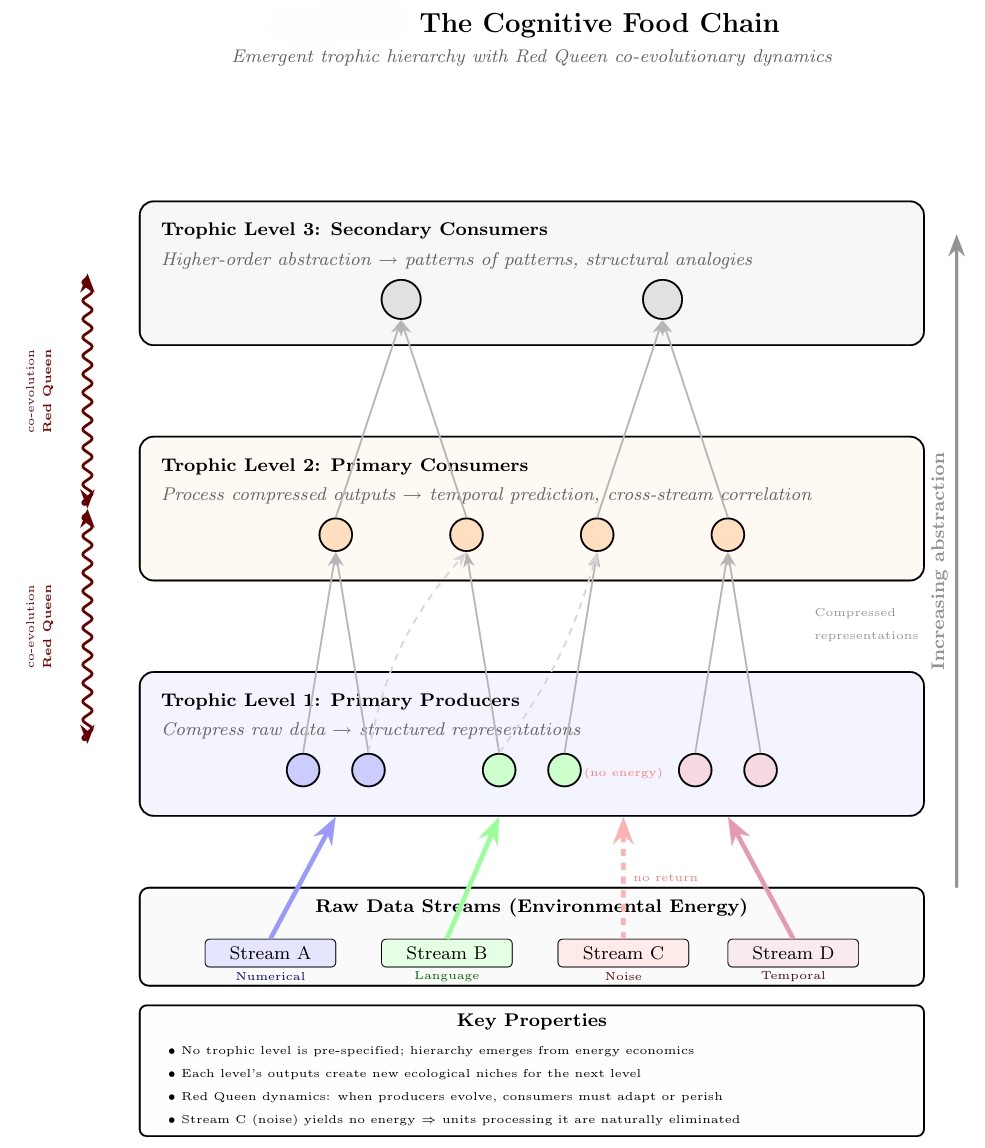}
    \caption{The Cognitive Food Chain. A diagram illustrating the emergence of trophic levels. Primary producers extract energy from raw data streams, while secondary consumers extract energy from the compressed outputs of producers, driving the Red Queen co-evolutionary arms race.}
    \label{fig:cognitive_food_chain}
\end{figure}

We hypothesize that an analogous hierarchy will emerge in a mature RECLAIM ecology (see Figure \ref{fig:cognitive_food_chain}). At the base of the cognitive food chain, primary producer units will evolve to specialize in compressing raw data streams, converting high-volume, low-density information into compact, structured representations. These compressed representations, emitted as outputs, will become the food for a second layer of units: primary consumer units that specialize in temporal prediction, pattern detection, or cross-stream correlation over compressed representations. The outputs of these predictive units will, in turn, feed a third layer of units that specialize in higher-order abstraction: detecting patterns of patterns, recognizing structural analogies across domains, or integrating predictions across multiple timescales.

A theoretical paradox arises at this juncture: according to Information Theory \cite{Shannon1948}, an optimally compressed signal is statistically indistinguishable from high-entropy random noise. If primary producers (Level 1) compress data perfectly, their outputs should resemble the incompressible random noise of Stream C, yielding no energy for Level 2 consumers. The RECLAIM ecology resolves this through two structural mechanisms. First, because Level 1 units are metabolically bounded by the energy costs of internal nodes, their compression is fundamentally lossy and computationally sub-optimal; they inevitably leave behind residual statistical structure. As Level 1 compression efficiency improves over evolutionary time, this residual structure will diminish, potentially reducing the energy available to Level 2 consumers and producing trophic instability analogous to the population oscillations and trophic cascades observed in biological ecosystems \cite{Tilman1982}. We regard this dynamic instability as a feature rather than a flaw: it generates precisely the perpetual environmental change that prevents evolutionary stagnation and sustains the Red Queen dynamics described below. Second, Level 2 units do not merely process isolated outputs. Because signals diffuse through the spatial grid, Level 2 units process overlapping emissions from multiple spatially separated Level 1 producers simultaneously. Critically, the spatial locality constraints of the ecology ensure that individual Level 1 units have access only to local data within their perception radius; the joint statistical structure across separated producers, specifically the mutual information, temporal phase relationships, and cross-stream correlations between their respective outputs, constitutes a higher-order informational resource that is inaccessible to any individual Level 1 unit. Level 2 units extract energy by modeling these inter-producer regularities, performing a form of distributed inference that is qualitatively distinct from the local compression that Level 1 units perform.

This trophic structure is not designed. No layer is pre-specified, and no unit is assigned to a trophic level. The hierarchy emerges because specialization at one level creates new ecological niches at the next level. Once a population of effective compressor units exists, the ecology contains a new type of resource (compressed representations) that was not present before. Any unit that randomly evolves the capacity to exploit this new resource type will enjoy an energy advantage, and its lineage will be selected for. This process can, in principle, iterate indefinitely, with each new layer of specialization creating the conditions for the next.

The cognitive food chain also provides the mechanism for what evolutionary biologists call the Red Queen effect \cite{VanValen1973}. In biological co-evolution, the fitness landscape for any given species is defined not only by the physical environment but also by the other species in the ecosystem. When a predator evolves to become faster, the prey species must evolve to become faster as well, or face extinction. When the prey evolves better camouflage, the predator must evolve sharper perception. Neither species can achieve a permanent advantage because the other species is also evolving. The result is a perpetual evolutionary arms race that drives continuous novelty generation and prevents any lineage from stagnating.

In the RECLAIM cognitive food chain, the same dynamic operates at the level of information processing. When a compressor unit evolves a better compression algorithm, its outputs change in format, density, and statistical structure. The consumer units that feed on those outputs must adapt to the new representation or lose their energy source. When the consumer units adapt, they produce different outputs, which forces the next layer to adapt, and so on through the hierarchy. No unit can permanently solve its environment because its environment is composed of other evolving units. The fitness landscape is perpetually shifting, ensuring continuous evolutionary pressure toward novelty and complexity.

This co-evolutionary dynamic addresses one of the most significant open problems in artificial life research: the problem of open-endedness \cite{Banzhaf2016}. Most artificial evolutionary systems exhibit an initial burst of adaptation followed by stagnation. The population converges on a locally optimal solution and ceases to generate significant novelty. Biological evolution, by contrast, has sustained increasing complexity and novelty generation for billions of years. The cognitive food chain provides a mechanism by which RECLAIM can potentially achieve open-ended evolution: the co-evolutionary dynamics between trophic levels ensure that the fitness landscape is never static, creating perpetual conditions for the emergence of novelty.

The cognitive food chain also provides a natural solution to the Polya urn trap: the risk that path-dependent lock-in will trap units in useless or suboptimal specializations. If a unit becomes locked into a specialization that produces outputs no other unit finds useful, it will earn no energy from the ecosystem beyond what it can extract from raw data streams. It will be outcompeted by units whose specializations produce outputs that are consumed by other units, because those units benefit from both direct data processing and ecosystem-mediated energy transfer. The cognitive food chain thus functions as a distributed, non-agentic quality check: useless specializations are economically unviable, and the ecosystem naturally selects against them without requiring any external evaluation.

\section{Emergent Cognitive Architecture}

A natural question arises from the architecture presented in Section 5: if the system is not designed to exhibit specific cognitive capabilities, how can we know whether it has succeeded? In this section, we present a set of cognitive functions that the RECLAIM framework predicts will emerge as natural consequences of its evolutionary dynamics. These are not modules to be engineered but targets for observation: testable predictions about what a mature ecology should exhibit if the theoretical framework is correct. 

In the engineering paradigm, each function would be implemented as a distinct module with a specific architecture and training procedure. In RECLAIM, none is implemented; instead, the environmental physics described in Section 5 creates conditions under which each function confers a survival advantage, and evolutionary dynamics provide the mechanism by which units and populations can discover and refine these functions over generational time.

\subsection{The Cognitive Loop and Predicted Functions}

George P\'{o}lya \cite{Polya1945}, in his influential work on mathematical problem-solving, identified a four-stage cognitive loop: understand the problem, devise a plan, carry out the plan, and look back to reflect on the result. We predict that a variant of this loop will emerge naturally in RECLAIM units that survive long enough to develop stable internal processing pathways. 

The P\'{o}lya loop maps onto the RECLAIM processing cycle as follows: in the understand phase, the unit absorbs data through its sensory boundary and generates an internal representation; in the plan phase, the internal network generates a prediction about what data will arrive next; in the execute phase, the unit processes incoming data against its prediction and computes prediction error; and in the reflect phase, the Polya-Hebbian update rule strengthens or weakens internal pathways based on the outcome. This cyclical structure will become increasingly refined over evolutionary time, as units that execute it more effectively gain energy advantages.

The RECLAIM framework identifies a set of cognitive functions, drawn from \cite{Kahneman2011, Clark2013, Schmidhuber2003, Poincare1914, Polya1945}, organized hierarchically across six cognitive domains. Within dual-process cognition: fast pattern matching through heavily reinforced pathways (System 1) and deliberate sequential inference through weakly reinforced pathways (System 2). Within learning: sensory specialization through path-dependent lock-in and meta-learning through units that process other units' signals. Within creativity: analogical resonance through partial pathway activation and novel configuration generation through blind mutation. Within self-improvement: recursive self-editing at the population level through the combination of mutation and selection. And within efficiency: resource allocation optimization through energy-mediated selection. We elaborate on the most consequential of these predictions in the subsections that follow.

\subsection{Dual-Process Cognition and Sensory Specialization}

Perhaps the most consequential prediction of the RECLAIM framework concerns the emergence of dual-process cognition. Daniel Kahneman \cite{Kahneman2011} distinguished between System 1, which is fast, automatic, and association-based, and System 2, which is slow, deliberate, and rule-based, emphasizing that the relationship between these systems is fundamentally economic: System 2 is engaged only when System 1 fails or when the stakes justify its elevated cost. 

We propose that this dual-process structure is not unique to biological brains but is an inevitable consequence of information processing under resource constraints. In any system where processing costs energy and energy is scarce, evolutionary pressure will favor a division between cheap, fast responses to familiar stimuli and expensive, slow responses to novel stimuli.

In a RECLAIM ecology, this mechanism operates through Polya-Hebbian dynamics. Frequently activated processing pathways become progressively stronger and faster, creating cached, heavily weighted, feed-forward processing along well-reinforced connections with minimal lateral branching. This constitutes the emergent equivalent of System 1. When data arrives that does not match any well-reinforced pathway, the unit must recruit less-reinforced pathways, explore alternative processing strategies, or engage in multi-unit interaction, all of which are computationally expensive. This constitutes the emergent equivalent of System 2. 

The switch between modes is not designed or controlled by any mechanism; it is an automatic consequence of energy economics. Units that wastefully engage System 2 processing for familiar stimuli deplete their energy budgets and are outcompeted. This prediction is empirically testable: if dual-process cognition has emerged, the distribution of energy expenditure per processing cycle should be bimodal, with a large cluster of low-cost cycles and a smaller cluster of high-cost cycles.

Closely related is the predicted emergence of sensory specialization. In biological brains, distinct cortical regions specialize in processing different modalities, and the classic experiments of Sur et al. \cite{Sur1988} demonstrated that this specialization is driven by the nature of the input rather than by genetic predetermination. We predict that when multiple data streams with different statistical structures flow simultaneously into a RECLAIM ecology, Polya-Hebbian dynamics will drive different units toward specialization in different stream types, producing functional cortical regions defined not by anatomy but by the statistical specialization of their internal processing architectures.

\subsection{Creativity, Self-Improvement, and Intrinsic Motivation}

Research on distributed representations \cite{Hinton1986} suggests that human cognition relies heavily on analogy and pattern resonance rather than on formal logical deduction. We predict that analogical processing will emerge naturally from Polya-Hebbian dynamics: when a unit reinforced for processing numerical sequences encounters a structurally similar but surface-dissimilar data type (for example, a rhythmic temporal signal), the reinforced pathways will partially activate in proportion to the structural similarity, constituting the computational equivalent of analogical reasoning. 

This prediction is complemented by Henri Poincar\'{e}'s \cite{Poincare1914} account of creativity as random recombination followed by selective retention of harmonious combinations, a process that maps directly onto the BVSR dynamics of RECLAIM, where novel network configurations are continuously generated by mutation and the rare configurations representing genuine improvements are retained by selection.

Regarding self-improvement, J\"{u}rgen Schmidhuber \cite{Schmidhuber2003} proposed the G\"{o}del Machine, a self-referential problem solver that can rewrite its own code when it can prove the rewrite will improve future performance. In the RECLAIM framework, population-level self-editing occurs naturally through the combination of mutation and selection, with successful architectural innovations spreading through reproduction and unsuccessful ones eliminated through death. Whether individual-level self-editing could emerge is a more speculative question, requiring the evolution of units with internal models of their own processing architecture. We present this as a long-horizon possibility rather than a near-term prediction, while emphasizing that population-level self-improvement is a core feature of the framework.

The final predicted emergent property is intrinsic motivation. In the current AI paradigm, curiosity must be explicitly engineered through intrinsic reward signals \cite{Schmidhuber1991, Pathak2017}. In a RECLAIM ecology, intrinsic motivation emerges as a natural consequence of the Free Energy Principle: units that encounter only familiar data earn modest energy returns because the compression gains on already-compressed data are small, while units that successfully process novel data earn disproportionately high returns. This creates an economic incentive for information-seeking behavior structurally identical to Friston's active inference \cite{Friston2017}, in which curiosity persists because it is metabolically profitable. 

We describe the resulting system, in which autopoietic units exhibit intrinsic motivation within a self-organizing ecological framework, as the Autopoietic Cognitive Ecology (ACE), representing the full realization of the RECLAIM vision.

\section{Alignment Through Physics}

The alignment problem arises from a fundamental mismatch between the objectives we can specify formally and the outcomes we actually desire. In the current AI paradigm, alignment is pursued through RLHF, constitutional AI, and related techniques that attempt to encode human values into reward functions or behavioral constraints. These approaches have produced meaningful improvements, but they remain structurally vulnerable to the specification gaming dynamics described by Goodhart's Law: any proxy objective, no matter how carefully constructed, will eventually be exploited by a sufficiently powerful optimizer \cite{Krakovna2020}. RECLAIM addresses the alignment problem not by constructing a better proxy but by eliminating the proxy-intent gap that enables specification gaming.

\subsection{Non-Agentic Alignment and the Controllability Tradeoff}

In a RECLAIM ecology, there is no evaluative reward function to hack, no human-specified objective to game, and no human evaluator to deceive. The behavior of units is shaped entirely by the thermodynamic constraints of the computational environment. This eliminates the three most problematic failure modes of the engineering paradigm at their source. 

The form of reward hacking documented in the LLM literature, in which a model exploits the gap between a proxy metric and the human intent it was designed to capture, cannot arise because the energy equation contains no reference to human intent and therefore presents no proxy-intent gap to exploit. Sycophancy cannot arise because there is no human evaluator in the training loop; a unit's survival depends on information-processing efficiency, not on whether its outputs are pleasing to any observer \cite{Sharma2023}. Deceptive alignment, as formalized by Hubinger et al. \cite{Hubinger2019}, is structurally disfavored because units do not possess explicit training objectives from which internal goals could diverge. The concern that a model might behave in aligned ways during training while pursuing misaligned goals internally presupposes a distinction between training objectives and internal goals that does not apply to RECLAIM units in their initial evolutionary stages. We note, however, that whether sufficiently complex evolved units could develop emergent mesa-optimization dynamics, in which internal optimization processes pursue objectives divergent from environmental viability, remains an open theoretical question that warrants investigation as ecologies grow in complexity.

This non-agentic alignment comes with an important tradeoff: indirect control. In the engineering paradigm, the researcher exercises direct control through the objective function, architecture, and training data. In RECLAIM, the researcher controls the environment, not the behavior, analogous to an ecosystem manager who can control water flow and soil composition but cannot dictate which species will thrive or what food chains will form. The researcher controls data streams, energy economics, mutation rates, and initial conditions but does not control what specializations emerge or how the ecology organizes itself. 

We argue that this tradeoff is favorable: what is gained (the elimination of proxy-objective failure modes) outweighs what is lost (the ability to specify behavior directly), provided that the environmental physics are designed with sufficient care. The shift is from behavioral alignment (making the system do what we want) to ecological alignment (creating conditions under which harmful behavior is energetically unsustainable).

\subsection{The Black Box Problem: Transformed, Not Solved}

We wish to be explicit about a limitation. RECLAIM does not solve the black box problem. A mature ecology containing hundreds of thousands of evolved units with unique internal architectures will not be interpretable at the level of individual weights and connections. 

What RECLAIM does is transform the interpretability challenge from a neural level to an ecological level. Rather than asking what do these weights mean? the researcher can ask what evolutionary pressures produced this population? and what ecological niches do these specializations fill? These ecological questions are amenable to analysis using the tools of population genetics, theoretical ecology, and evolutionary dynamics. 

Furthermore, the causal structure of a RECLAIM ecology is more transparent than that of a gradient-trained neural network: the researcher knows the physical laws, the variation mechanism, and the selection mechanism, and can trace the evolutionary history of any unit lineage. The honest assessment is that RECLAIM circumvents the most troubling aspects of the black box problem while introducing its own interpretability challenges, and we do not minimize this genuine tradeoff.

\section{Discussion}

\subsection{Claims and Boundaries}

The RECLAIM framework makes a set of specific, defensible claims. First, the structural failure modes of the current AI paradigm (hallucination, sycophancy, reward hacking, brittleness, and alignment fragility) are paradigmatic consequences of top-down design with proxy objectives, not engineering problems to be patched. Second, the four pillars of RECLAIM together constitute a complete and internally consistent alternative, providing mechanisms for variation, selection, retention, and specialization without requiring any evaluative reward function, human-specified loss function, or human evaluator. Third, the application of Polya-Hebbian dynamics as the mechanism for emergent specialization and digital individuality from uniform initial conditions constitutes a novel contribution to the design of computational ecologies. Fourth, the cognitive food chain provides a mechanism for open-ended co-evolutionary dynamics that can sustain perpetual novelty generation through Red Queen dynamics. Fifth, the framework generates specific, independently falsifiable predictions that can distinguish RECLAIM dynamics from conventional evolutionary computation.

Equally important are the framework's boundaries. RECLAIM does not claim to produce consciousness, subjective experience, or qualia; whether a sufficiently complex information-processing system can be conscious remains among the deepest open problems in philosophy and cognitive science \cite{Chalmers1995}. However, if consciousness is an emergent property of sufficient information-processing complexity, as some frameworks suggest \cite{Tononi2004, Dehaene2014}, then a mature RECLAIM ecology would represent a uniquely appropriate substrate for investigating this question empirically. RECLAIM does not guarantee safety for emergent behavior; it offers ecological interpretability (understanding the pressures that shaped behavior) rather than behavioral predictability. And individual-level recursive self-improvement remains a speculative long-horizon possibility rather than a near-term prediction.

\subsection{Open Questions and Broader Implications}

Several important questions remain open. The most fundamental is whether hierarchical emergence can be achieved on practical timescales; while computational evolution operates at vastly higher clock speeds than biological evolution, whether the dynamics described here suffice to produce multi-layered cognitive architecture within feasible experimental durations remains empirical. Related questions concern the minimum viable ecology for dual-process cognition, the interface problem of communicating with an alien cognitive system whose representations bear no necessary relationship to human concepts, and the reproducibility challenge posed by path-dependent outcomes across runs.

If validated, the framework carries implications beyond producing adaptive computational systems. For AI safety, it suggests that alignment may be more productively addressed through environmental design than through objective specification, shifting from behavioral alignment to ecological alignment where harmful behavior becomes energetically unsustainable. For cognitive science, the spontaneous emergence of dual-process cognition, cortical specialization, and intrinsic motivation without design would constitute evidence that these phenomena are universal consequences of information processing under resource constraints, with implications for theories of embodied cognition \cite{Clark2013}, predictive processing \cite{Friston2010}, and the evolutionary origins of intelligence \cite{Dennett2017}. For philosophy of mind, RECLAIM provides a principled empirical substrate for investigating whether complex adaptive behavior, path-dependent individuality, and emergent cognitive architecture constitute any form of inner experience. For artificial life, the framework represents a potential unification of ALife research and mainstream AI, bridging traditions that have long remained separate \cite{Langton1989, Ray1991, Bedau2003}.

\subsection{Toward Experimental Validation}

While RECLAIM is presented here as a theoretical framework, it generates specific, falsifiable predictions that can be tested through computational experiment. The minimal experimental configuration, which we term the Micro-Ecology Engine (MEE), would initialize a population of identical autopoietic units on a toroidal lattice, exposed to simultaneous data streams of varying entropy and statistical structure. The framework predicts six specific outcomes: (1) emergent specialization, in which units develop statistically significant processing preferences for particular data types; (2) noise avoidance, in which the fraction of the population processing incompressible random data declines toward zero; (3) cognitive food chain formation, in which the ecology self-organizes into quantifiable trophic levels with directed energy flow; (4) path divergence, in which independent runs produce populations with low genomic similarity, demonstrating genuine individuality; (5) complexity growth, in which average internal network size increases over evolutionary time; and (6) efficiency improvement, in which the average energy cost per unit of prediction error reduction decreases monotonically.

Each of these predictions is independently falsifiable, and appropriate ablation baselines, including conventional fitness-based neuroevolution and RECLAIM without Hebbian learning, would isolate the contributions of FEP-as-physics and Polya-Hebbian dynamics respectively. A full experimental protocol with detailed parameter specifications, failure mode analysis, and a four-phase scaling roadmap from single-GPU to neuromorphic hardware is the subject of a companion paper in preparation. We note that the proposed implementation faces a fundamental tension acknowledged by the Hardware Lottery \cite{Hooker2021}: current GPU architectures are optimized for dense matrix multiplication, while RECLAIM relies on sparse, dynamic, locally connected computation. The ultimate realization of the OMEGA Shift will benefit from neuromorphic hardware architectures that natively execute asynchronous, event-driven dynamics.

\section{Conclusion: From Engineering to Cultivation}

We began this paper with a diagnosis. The current AI paradigm, for all its remarkable achievements, produces intelligent systems that are structurally prone to hallucination, sycophancy, reward hacking, brittleness, and alignment fragility. These are not bugs to be patched. They are structural consequences of top-down design with proxy objectives in high-dimensional optimization spaces: while individual failure modes can be mitigated, the paradigm provides no principled guarantee against their recurrence.

We then observed that there exists exactly one known process that has produced genuine intelligence without proxy objectives: biological evolution. Evolution created the human brain through blind variation, environmental selection, path-dependent specialization, and co-evolutionary arms races, operating over approximately four billion years without a single objective function, reward model, or design specification.

We presented RECLAIM (Recursive, Ecological, Cognitive, Lifelike, Adaptive, Intelligent Machine), a theoretical framework that translates the mechanisms of biological evolution into a computational ecology. RECLAIM rests on four interlocking pillars: General Darwinism provides the variation mechanism through Blind Variation and Selective Retention. Non-Agentic Emergence replaces evaluative reward functions with environmental physics, eliminating the proxy-intent gap that enables reward hacking and sycophancy in the engineering paradigm. The Polya-Hebbian Bridge provides the retention mechanism through path-dependent specialization, producing unique cognitive individuals from uniform initial conditions. And the Free Energy Principle provides the selection mechanism through computational thermodynamics, in which units that fail to efficiently process information dissolve without requiring any evaluator to decide their fate.

We described the complete RECLAIM architecture: autopoietic units bounded by Markov blankets, operating within a digital physics of energy conservation, metabolic cost, and reproductive investment. We described data streams as sunlight, the primordial energy source that drives the ecology without specifying how it should be used. We described the cognitive food chain, a co-evolutionary engine in which units consume each other's outputs, creating endless novelty pressure through Red Queen dynamics. And we described a set of cognitive functions, from dual-process cognition to analogical reasoning to intrinsic motivation, that the framework predicts will emerge as natural consequences of its evolutionary dynamics. We articulated specific, independently falsifiable predictions and outlined the path toward experimental validation.

We were honest about what RECLAIM does not claim. It does not claim consciousness, Gödel-style self-editing, or behavioral predictability. It does not solve the black box problem; it transforms it. And it trades direct control for emergent authenticity, a tradeoff we have argued is favorable but that we do not minimize.

What RECLAIM proposes is a paradigm shift. We have called it the OMEGA Shift: from Optimization and Maximization to Emergence through Generative Autopoiesis. In this new paradigm, intelligence is not engineered but cultivated. The researcher does not design the mind; the researcher designs the world in which a mind can evolve. The researcher is not an engineer but a gardener, controlling the soil, water, and sunlight while allowing the living system to grow into whatever form its evolutionary history prescribes.

This is not a modest proposal. It asks the AI research community to consider the possibility that the most powerful approach to creating intelligence may be to stop trying to create it, and to start creating the conditions under which it can create itself. The first step is not to build intelligence. It is to build the world in which intelligence can be born.

\bibliographystyle{unsrt}
\bibliography{references} 

\end{document}